\documentclass[sigconf]{acmart}

\usepackage{graphicx}
\usepackage{amsmath}
\usepackage{booktabs}
\usepackage{subfloat}
\usepackage{subcaption}
\usepackage{multirow}
\usepackage{wrapfig}
\usepackage[T1]{fontenc}
\usepackage[ruled, linesnumbered]{algorithm2e}
\usepackage{xcolor}
\usepackage{colortbl}
\usepackage[normalem]{ulem}
\usepackage{enumitem}

\copyrightyear{2023}
\acmYear{2023}
\setcopyright{acmlicensed}\acmConference[MM '23]{Proceedings of the 31st
ACM International Conference on Multimedia}{October 29-November 3,
2023}{Ottawa, ON, Canada}
\acmBooktitle{Proceedings of the 31st ACM International Conference on
Multimedia (MM '23), October 29-November 3, 2023, Ottawa, ON, Canada}
\acmPrice{15.00}
\acmDOI{10.1145/3581783.3612318}
\acmISBN{979-8-4007-0108-5/23/10}

\begin{document}

\title{Parameter Exchange for Robust Dynamic Domain Generalization}
\author{Luojun Lin}
\affiliation{%
  \institution{Fuzhou University}
  \city{Fuzhou}
  \country{China}}
\email{linluojun2009@126.com}
\orcid{0000-0002-1141-2487}

\author{Zhifeng Shen}
\affiliation{%
  \institution{Fuzhou University}
  \city{Fuzhou}
  \country{China}}
\email{shen_zhifeng@outlook.com}
\orcid{0000-0002-2215-4479}

\author{Zhishu Sun}
\affiliation{%
  \institution{Fuzhou University}
  \city{Fuzhou}
  \country{China}}
\email{sunzhishu2013@163.com}
\orcid{0000-0002-0832-1509}

\author{Yuanlong Yu}
\authornote{Y. Yu and W. Chen are corresponding authors.}
\affiliation{%
  \institution{Fuzhou University}
  \city{Fuzhou}
  \country{China}}
\email{yu.yuanlong@fzu.edu.cn}
\orcid{0000-0002-2112-6214}

\author{Lei Zhang}
\affiliation{%
  \institution{Chongqing University}
  \city{Chongqing}
  \country{China}}
\email{leizhang@cqu.edu.cn}
\orcid{0000-0002-5305-8543}

\author{Weijie Chen}
\authornotemark[1]
\affiliation{%
  \institution{Zhejiang University \& Hikvision Research Institute}
  \city{Hangzhou}
  \country{China}}
\email{chenweijie@zju.edu.cn}
\orcid{0000-0001-5508-473X}

\renewcommand{\shortauthors}{Luojun Lin et al.}

\begin{abstract}
Agnostic domain shift is the main reason of model degradation on the unknown target domains, which brings an urgent need to develop \emph{Domain Generalization} (DG). Recent advances at DG use dynamic networks to achieve training-free adaptation on the unknown target domains, termed \emph{Dynamic Domain Generalization} (DDG), which compensates for the lack of self-adaptability in static models with fixed weights. The parameters of dynamic networks can be decoupled into a static and a dynamic component, which are designed to learn domain-invariant and domain-specific features, respectively. Based on the existing arts, in this work, we try to push the limits of DDG by disentangling the static and dynamic components more thoroughly from an optimization perspective. Our main consideration is that we can enable the static component to learn domain-invariant features more comprehensively by augmenting the domain-specific information. As a result, the more comprehensive domain-invariant features learned by the static component can then enforce the dynamic component to focus more on learning adaptive domain-specific features. To this end, we propose a simple yet effective \emph{Parameter Exchange} (PE) method to perturb the combination between the static and dynamic components. We optimize the model using the gradients from both the perturbed and non-perturbed feed-forward jointly to implicitly achieve the aforementioned disentanglement. In this way, the two components can be optimized in a mutually-beneficial manner, which can resist the agnostic domain shifts and improve the self-adaptability on the unknown target domain. Extensive experiments show that PE can be easily plugged into existing dynamic networks to improve their generalization ability without bells and whistles.
\end{abstract}



\maketitle


\section{Introduction}
Deep learning has achieved remarkable success in various vision tasks \cite{resnet,chen9981099}. Currently, most deep models are trained based on the \emph{i.i.d} assumption that the training and testing data are independent identically distributed, but it does not always hold in reality due to the agnostic domain shifts between training and testing data. Agnostic domain shifts usually cause deep models to suffer from drastic performance degradation when tested, especially for data from unknown domains. This results in an urgent need for \textbf{D}omain \textbf{G}eneralization (DG), which is a recently-popular research topic that aims to improve the cross-domain generalization ability of deep models against agnostic domain shifts.

\begin{figure}[t]
     \centering
     \begin{subfigure}[b]{0.49\columnwidth}
         \centering
         \includegraphics[width=\textwidth]{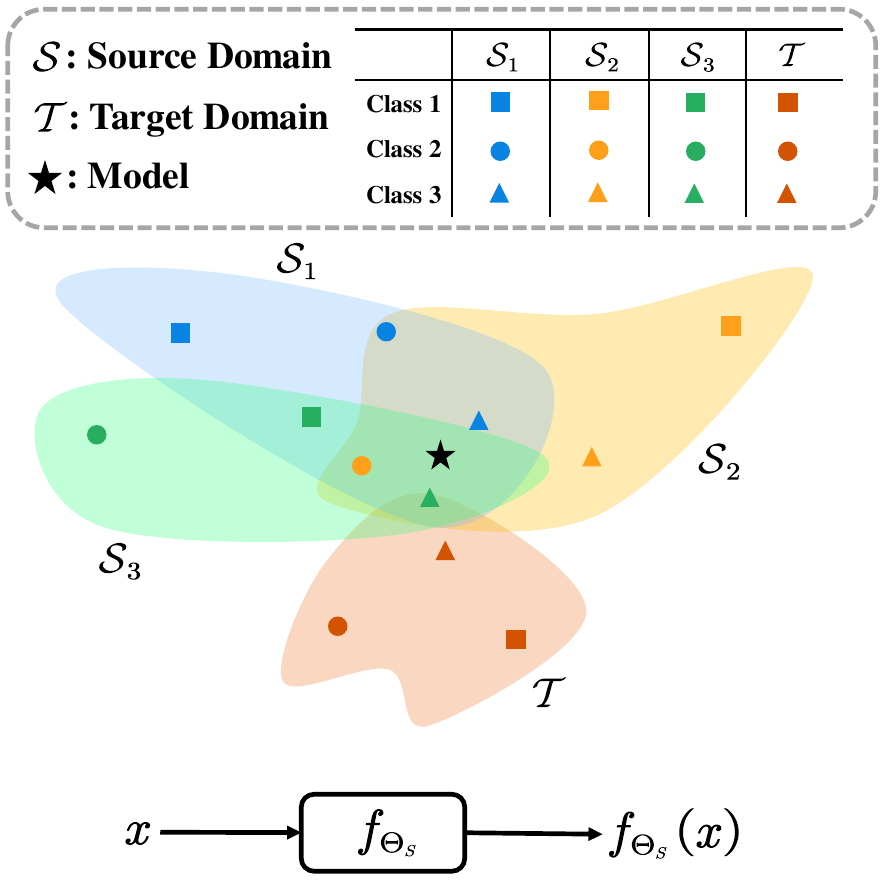}
         \caption{Static DG}
         \label{fig:static_dg}
     \end{subfigure}
     \hfill
     \begin{subfigure}[b]{0.49\columnwidth}
         \centering
         \includegraphics[width=\textwidth]{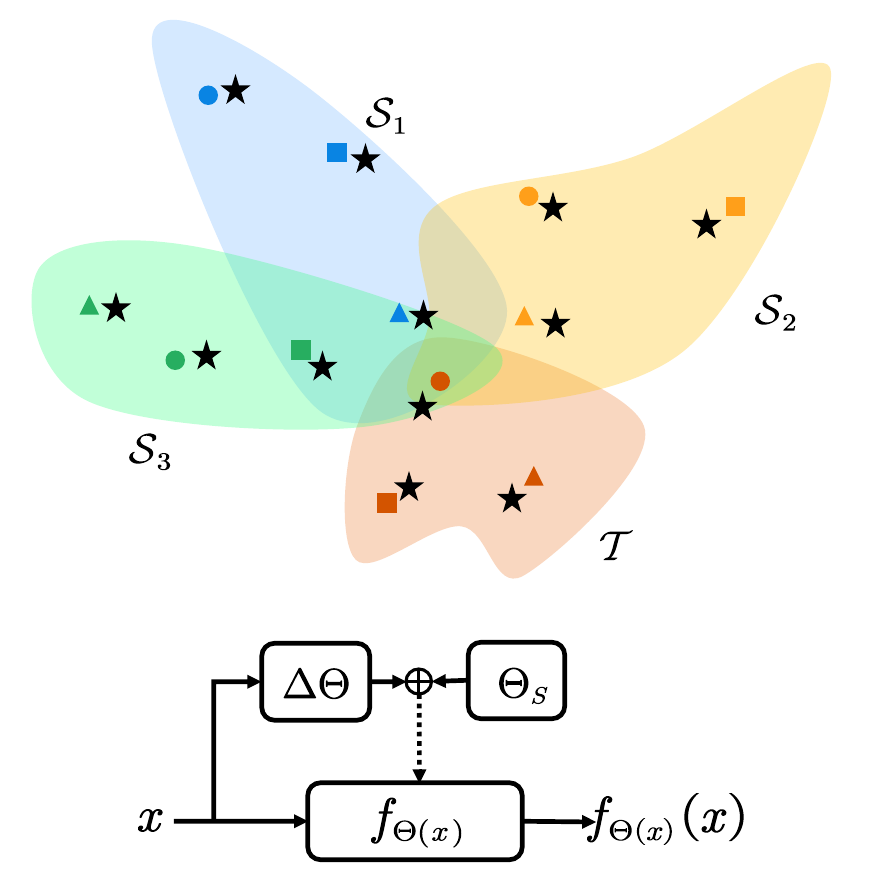}
         \caption{Dynamic DG}
         \label{fig:dynamic_dg}
     \end{subfigure}\\
        \vspace{-0.3cm}
        \caption{Static DG \emph{vs.} dynamic DG~(DDG). The static DG shares a common model among different domains; whilst the dynamic DG adjusts the model for each instance and thus making it easier to adapt to unknown target domain.}
        \label{fig:dynamic_vs_static}
         \vspace{-0.4cm}
\end{figure}

DG is usually formulated to train a deep model on multiple source domains and evaluate it on an unknown target domain. Most existing arts aim to obtain a robust static model by projecting multiple source domains into a common distribution, namely domain-invariant space~\cite{c-dann, sicilia2021dann, matsuura2020domain, ddaig}, where the domain-specific information is neglected, resulting in an adaptation failure to the unknown target domains (see Figure \ref{fig:dynamic_vs_static}(a)). Fortunately, the limitation of static models has been alleviated by \emph{Dynamic Domain Generalization (DDG)}~\cite{sun2022ddg}, which develops a dynamic network to achieve training-free adaptation on the unknown target domains, as shown in Figure \ref{fig:dynamic_vs_static}(b). By decoupling the network parameters into a \emph{static component} and a \emph{dynamic component}, DDG allows the static component to learn \emph{domain-invariant} features shared across instances, while the dynamic component adapts to each instance on the fly, learning \emph{domain-specific} features. In this way, DDG compensates for the lack of self-adaptability in static models, which rely solely on domain-invariant features.

Previous DG studies based on static networks~\cite{chattopadhyay2020learning, bui2021exploiting} have demonstrated that the domain-invariant and domain-specific features can complement each other in a mutually beneficial way to learn robust representations. However, there still lacks effective constraints to guide the optimization of the dynamic network towards this direction, which is the main limitation of DDG. In this paper, we consider to push the limits of DDG on the basis of existing arts, by disentangling the domain-invariant and domain-specific features more thoroughly from the perspective of optimization. The design objective of disentanglement is to enforce the static component of the dynamic network to learn domain-invariant features more comprehensively, which, in turn, would facilitate the dynamic component to focus more on learning adaptive domain-specific features.

To this end, we propose a simple yet effective \emph{Parameter Exchange (PE)} method to implicitly achieve the aforementioned disentanglement, which mainly changes the combination between the static and dynamic components by perturbing the domain-specific information encoded in the dynamic component. We implement PE that updates the model using the gradients from both the perturbed and non-perturbed feed-forward jointly. By doing so, the static component is enforced to be invariant to the perturbation from the dynamic component, as guided by the gradient from the perturbed feed-forward. Meanwhile, the gradient from the non-perturbed feed-forward prevents unstable training and promotes the dynamic component to be adaptive to the instance, under the cooperation with the robust domain-invariant features from the static component. In this way, the static and dynamic components are facilitated and decoupled mutually to promote the disentanglement of the domain-invariant and domain-specific features, ultimately improving the generalization ability of DDG. Specifically, we also design two variants of PE, including \emph{cross-instance PE} that shuffles the dynamic parameters across different instances, and \emph{cross-kernel PE} that reassigns them across different kernels, as shown in Figure \ref{fig:cross_attetion}.

Our method is evaluated on multiple standard and popular DG benchmarks, including PACS ~\cite{pacs}, Office-Home~\cite{venkateswara2017OfficeHome}, VLCS~\cite{vlcs}, TerraIncognita~\cite{terraincognita}, and DomainNet~\cite{DomainNet}, which vary in size, scenarios, categories and domain shifts. The experimental results show that our method can further improve the performance based on the strong baselines provided by DDG, which sufficiently demonstrates the effectiveness of the proposed PE method. In summary, we have found a new way to study DG from the perspective of disentangling dynamic networks, which not only promotes the research of DG, but also has significance for the field of dynamic networks. 

Overall, our contribution can be summarized as follows:
\begin{itemize}[leftmargin=12pt, topsep=2pt, itemsep=0pt]
    \item Based on the existing art~\cite{sun2022ddg}, we propose a simple yet effective \emph{Parameter Exchange} method that aims to disentangle the domain-invariant and -specific features more thoroughly by perturbing the combination between the static and dynamic components.
    \item Specifically, we develop two different PE manners: \emph{cross-instance PE} that exchanges the parameters of dynamic component across different instances, and \emph{cross-kernel PE} that reassigns them across different kernels. Also, to facilitate the feature disentanglement, the model is updated using the gradients from both the perturbed and non-perturbed feed-forward jointly.
    \item Extensive experimental results show that PE is a \emph{plug-and-play} method that can be easily applied to existing dynamic networks, such as DDG~\cite{sun2022ddg}, DRT~\cite{li2021dynamictransfer}, and ODConv~\cite{li2022omnidimensional}, to significantly improve their generalization performance on different DG benchmarks. In particular, our method integrated with ODConv achieves state-of-the-art performance on multiple DG benchmarks, indicating the effectiveness of PE in advancing the dynamic domain generalization. Our code is available at \url{https://github.com/MetaVisionLab/PE}.
\end{itemize}
\vspace{-\topsep}

\section{Related Work}
\noindent\paragraph{\textbf{Conventional Domain Generalization}} 
Conventional DG aims to train a robust static model on multiple source domains that performs well on unknown target domains. Most existing DG methods can be divided into several categories, including data/feature augmentation, representation learning, meta-learning, model ensemble, etc. In the early stages, many researchers hold the view that the deep features should be invariant in both the source domains and the unseen target domain, leading to a core idea that minimizes the representation discrepancy among the source domains \cite{c-dann, sicilia2021dann, motiian2017unifiedDG, li2018adversarialfeatureDG, li2020domain, wang2021respecting, pacs, ding2017deep, chattopadhyay2020learning, chen2021style}.
Meta-learning is also used in DG approaches~\cite{balaji2018metareg, dou2019masf, du2020metalearningVIB, zhao2021metalearningpersonreindentification}, which is mainly built on the learning paradigm of \cite{li2018metalearningDG}, where the source domains are divided into meta-source and meta-target domains, and the learning objective is to minimize the test error on the meta-target domain using the model updated by the meta-source domain. Another line of DG approaches are involved with augmentation techniques, which aims to enrich the style information whilst leaving the image category unchanged. Most of augmentation-based methods is performed on data-level by synthesizing newly-stylized images~\cite{ddaig, zhou2020l2a_ot, xu2021fact, lin2021semi, adapt2023xie}, or on feature-level by randomly mixing the feature statistics that contains style information~\cite{mixstyle, li2021uncertainty}. Such data/feature augmentation serves as a regularization term to prevent the model from over-fitting, which has become a main direction in solving DG. Our approach can be categorized as a special case of augmentation-based methods. While many augmentation-based DG techniques rely on data augmentation or feature augmentation, our method stands out as the first to introduce parameter augmentation tailored for dynamic networks, presenting a novel avenue for the DG community.

\noindent\paragraph{\textbf{Dynamic Domain Generalization}}
In most existing DG methods, the network parameters are fixed once finishing the training on source domains, which may fail to generalize to agnostic target domains due to lack of adaptability. Recent advances have demonstrated a new insight for DG by developing a new domain generalization paradigm that allows deep models to be fit to the unknown target domain at test time~\cite{sun2020test,iwasawa2021test,zhao2022test}. However, test-time DG always requires additional computational overhead for test-time training, which is impractical to apply to edge devices.
A natural solution involves using a dynamic network to adjust the network parameters for different inputs~\cite{lin2019attribute}. Hu et al.~\cite{hu2022domain} exploit dynamic convolution to solve the generalized medical image segmentation task, where the filter parameters are generated by a multilayer perceptron (MLP). However, the growing number of parameters would increase the risk of overfitting. Some studies have been developed to solve this problem by squeezing the dynamic parameters into a low-dimensional space~\cite{wang2022domain, sun2022ddg}. Sun et al.~\cite{sun2022ddg} decomposes the network parameters into a dynamic component and a static component to reduce the dynamic parameter space and achieve dynamic domain generalization (DDG).
Based on the existing art~\cite{sun2022ddg}, this paper devises a new optimization constraint on the dynamic network to achieve feature disentanglement, which can finally strengthen DDG.

\section{Methodology}
\subsection{Recap of Dynamic Domain Generalization}
Before introducing our method, we first briefly recap dynamic domain generalization (DDG) from the perspectives of task formulation and network design.

\noindent\paragraph{\textbf{Task Formulation}}
For DG task, there are multiple source domains with a total of $N_S$ samples available for training. The training set is represented as $\mathcal{D}_{train}=\{\mathcal{X}_S,  \mathcal{Y}_S\}$, where $\mathcal{X}_S, \mathcal{Y}_S$ represent the image set and label set, respectively. The goal of conventional DG is to learn a robust static mapping $f_{\mathbf{\Theta}_s}:\mathcal{X}_S \rightarrow \mathcal{Y}_S$ to generalize to the unknown target domain. In contrast, the objective of DDG is to learn a dynamic mapping $f_{\mathbf{\Theta}(x)}:\mathcal{X}_S \rightarrow \mathcal{Y}_S$ that adapts to each instance, where the instance-aware adaptability can enhance its generalizability to the unknown target domain.

\begin{figure}[t]
     \centering
     \begin{subfigure}[b]{0.49\columnwidth}
         \centering
         \includegraphics[width=\columnwidth]{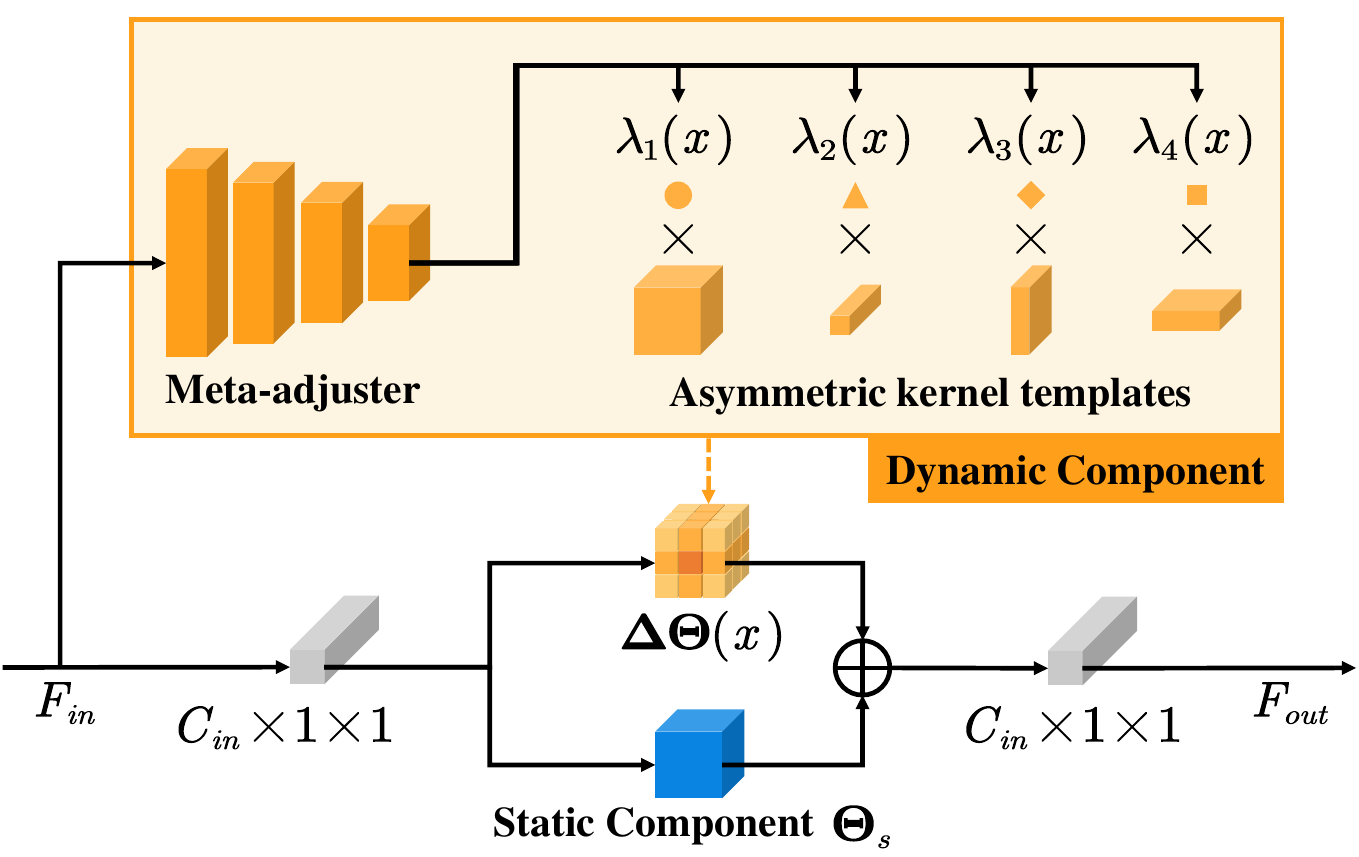}
         \caption{}
         \label{fig:block}
     \end{subfigure}
     \begin{subfigure}[b]{0.49\columnwidth}
         \centering
         \includegraphics[width=\columnwidth]{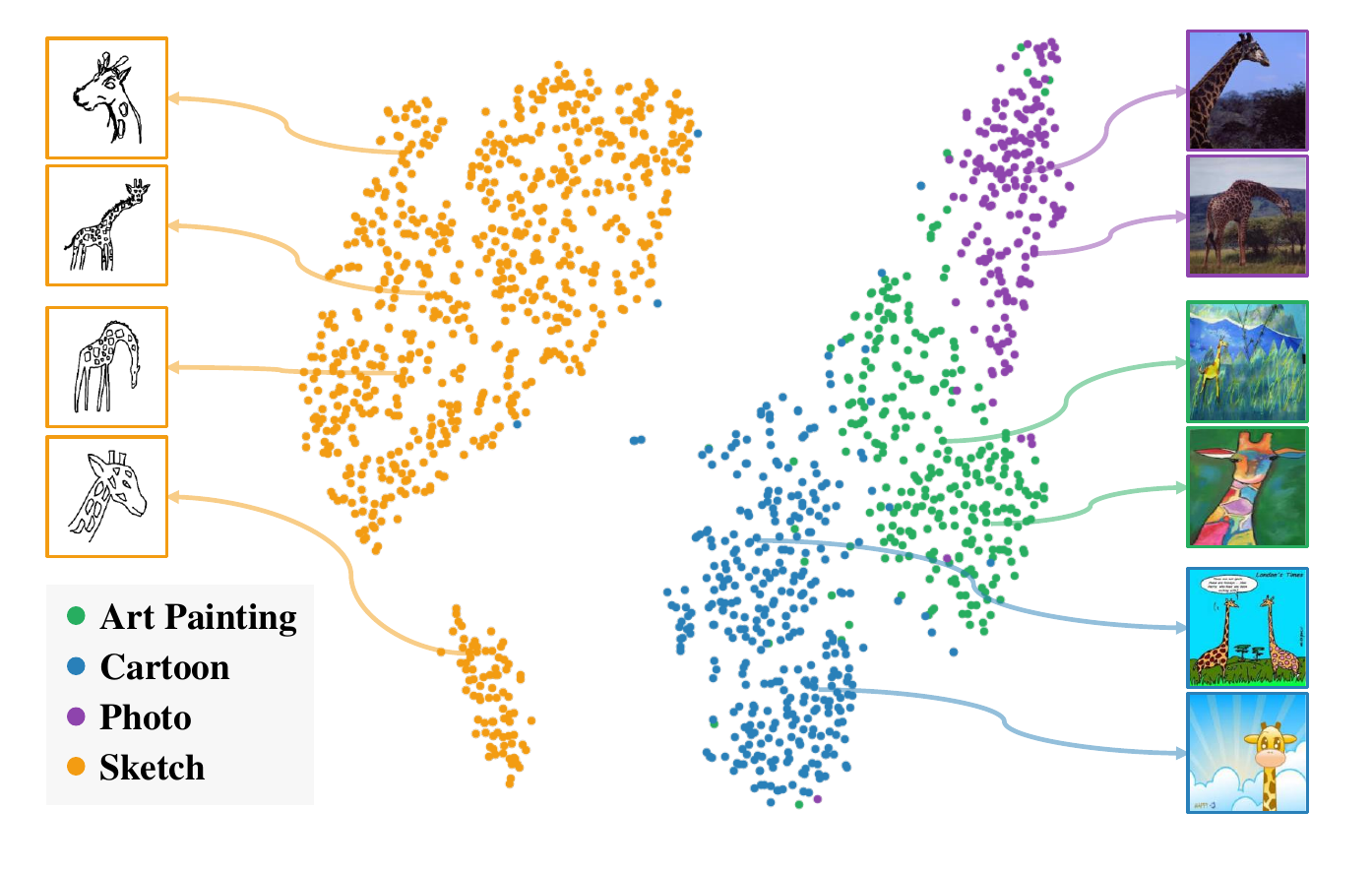}
         \caption{}
         \label{fig:coefftsne}
     \end{subfigure}\\
        \vspace{-0.3cm}
        \caption{(a) The basic block of the dynamic network in DDG consists of a static and a dynamic component. The dynamic component is a learnable linear combination between asymmetric kernel templates and dynamic coefficients generated by the meta-adjuster. The four asymmetric kernel templates are varied with size of $C_{in}\times K\times K$, $C_{in}\times 1\times 1$, $C_{in}\times K\times 1$ and $C_{in}\times 1\times K$. Note that the output channel $C_{out}$ is neglected here for simplicity.
        (b) t-SNE~\cite{van2008visualizing} visualization of dynamic coefficients by taking different domain data from PACS ~\cite{pacs} as input. The visualization of the dynamic coefficients shows a strong correlation with the domain labels, suggesting that the dynamic component can effectively capture domain or style information without requiring explicit supervision from domain labels.  Best viewed in colors.}
        \label{fig:block_and_coefftsne}
        \vspace{-0.2cm}
\end{figure}

\begin{figure*}[t]
     \centering
     \begin{subfigure}[b]{0.45\textwidth}
         \centering
         \includegraphics[width=\textwidth]{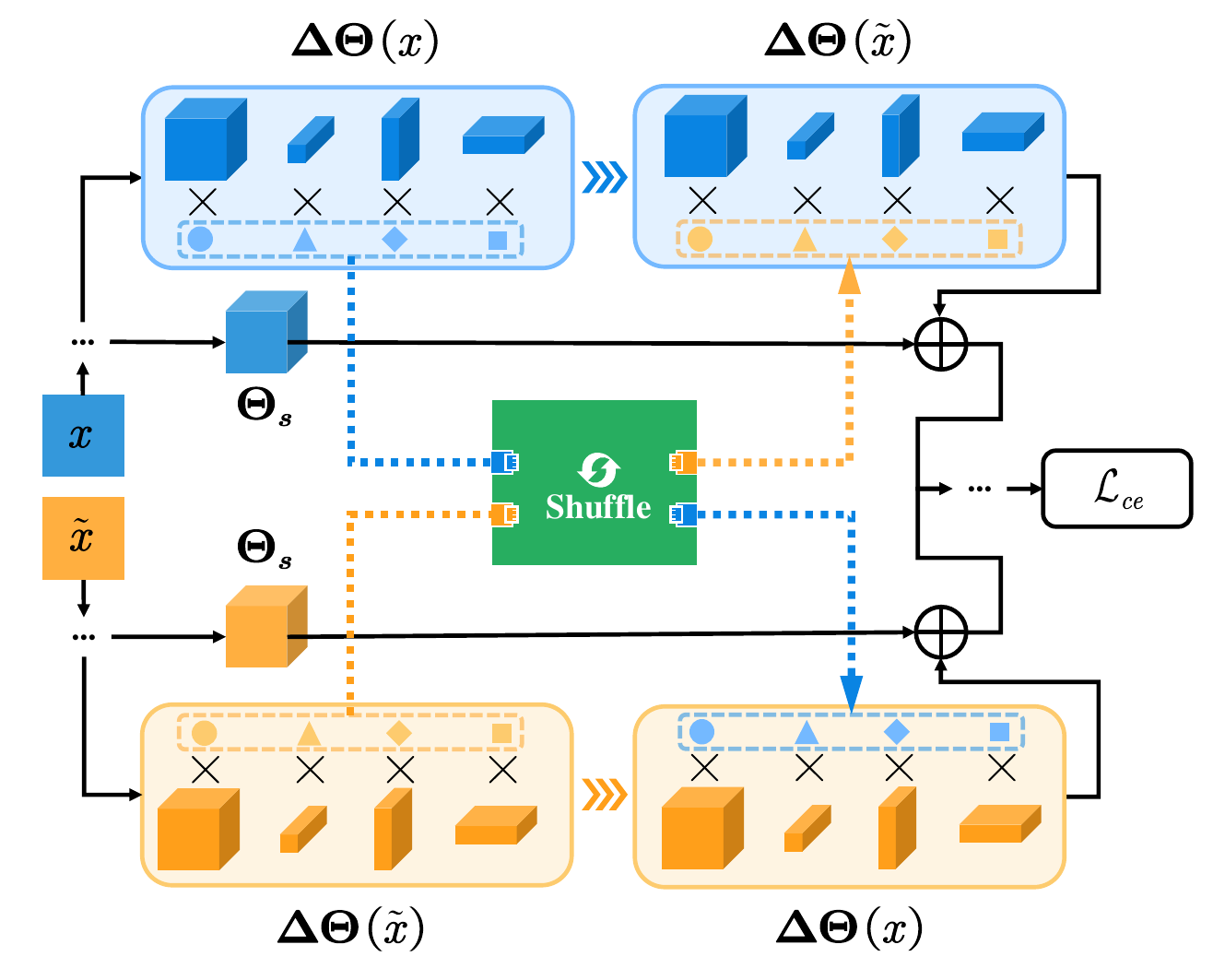}
         \caption{Cross-instance parameter exchange}
         \label{fig:cross_sample}
     \end{subfigure}
     \hspace*{2em}
     \begin{subfigure}[b]{0.45\textwidth}
         \centering
         \includegraphics[width=\textwidth]{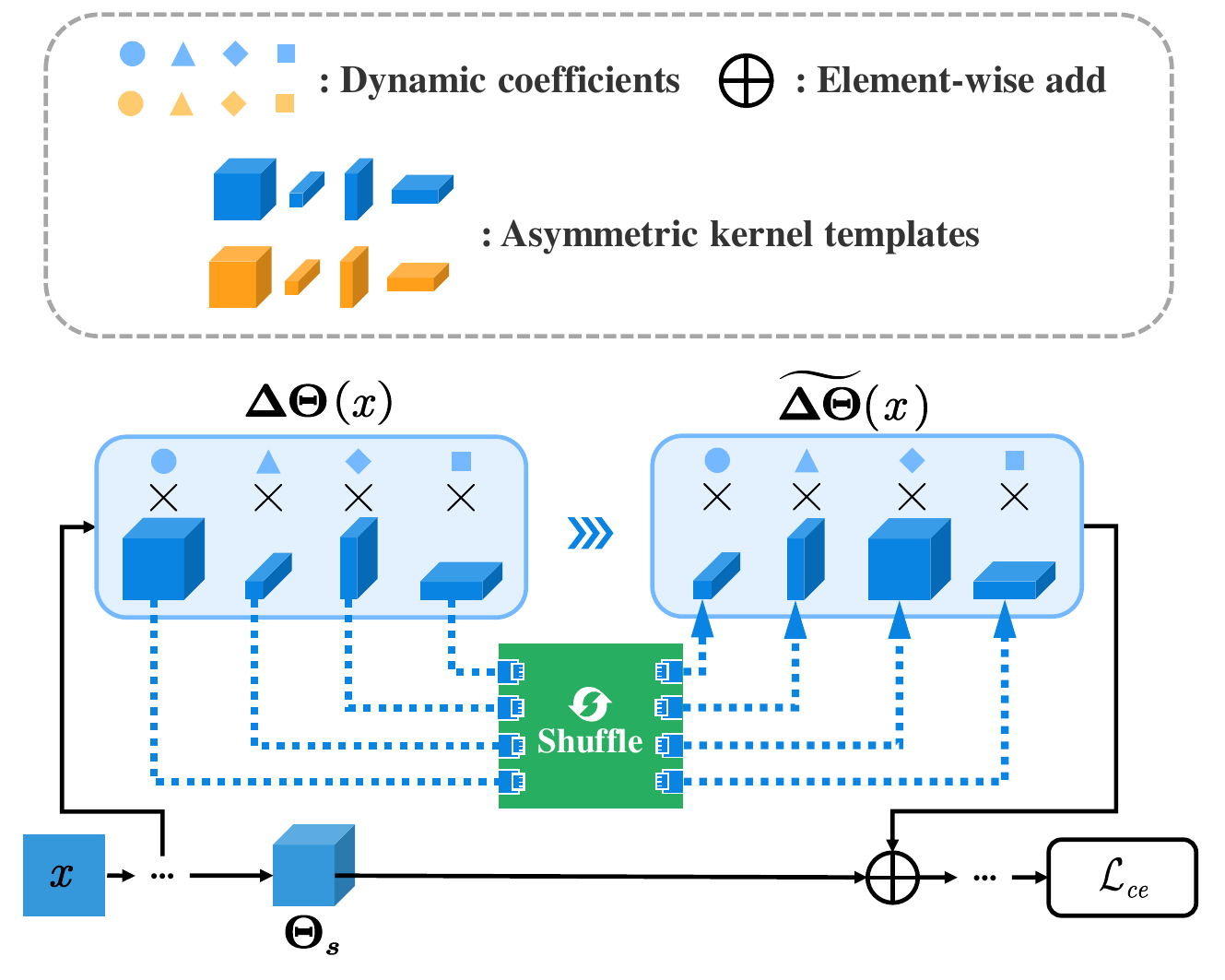}
         \caption{Cross-kernel parameter exchange}
         \label{fig:cross_kernel}
     \end{subfigure}\\
        \vspace{-0.3cm}
        \caption{Pipeline of parameter exchange (PE), including cross-instance PE and cross-kernel PE.}
        \label{fig:cross_attetion}
\end{figure*}

\noindent\paragraph{\textbf{Network Design}}
In DDG, a special designed dynamic network is employed to achieve instance-level adaptability~\cite{sun2022ddg}. As shown in Figure \ref{fig:block}, the network parameters are a function of the given input $x$, represented as $\boldsymbol{\Theta}(x)$, which can be divided into a static component $\boldsymbol{\Theta}_s$ and a dynamic component $\boldsymbol{\Delta\Theta}(x)$:
\begin{equation}
\boldsymbol{\Theta}(x) = \boldsymbol{\Theta}_s + \boldsymbol{\Delta\Theta}(x).
\label{eq:ddg_all}
\end{equation}
The static component is shared across different instances, while the dynamic component is adapted to each instance.
In particular, the dynamic component can be decomposed into a linear combination of several asymmetric kernel templates with low-dimensional dynamic coefficients:
\begin{equation}
\boldsymbol{\Delta\Theta}(x) = \sum_{m=1}^{M}{\lambda_m(x) \boldsymbol{\Phi}_m}, 
\label{eq:ddg_dy}
\end{equation}
where $M$ is a hyper-parameter specifying the number of kernel templates ($M=4$ by default). $\lambda_m(x)$ represents the $m$-th element in the dynamic coefficient set $\{\lambda_m(x)\}_{m=1}^{M}$, which practically is a probability distribution generated by the meta-adjuster $\lambda(\cdot)$ conditioned on the input $x$. $\boldsymbol{\Phi}_m$ represents the $m$-th kernel template in the asymmetric kernel template set $\{\boldsymbol{\Phi}_m\}_{m=1}^M$, in which each kernel template is designed with asymmetric shapes along the channel, height and width dimensions, in order to strengthen the central criss-cross positions of kernels that can prevent model collapse. These asymmetric kernel templates can be mapped to the same size via zero padding and then integrated into a dense $C_{in} \times K\times K$ kernel, namely dynamic component.

\subsection{Training with Parameter Exchange}

\noindent\paragraph{\textbf{Rethinking Dynamic Domain Generalization}}
Before presenting the proposed method, let us revisit how the static and dynamic components interact in DDG.
The static component is shared among different instances and therefore captures the task-oriented features shared across domains, also known as domain-invariant features. In contrast, the dynamic component is adapted to each instance and therefore captures instance-level task-oriented features, also known as domain-specific features since each instance can be regarded as an individual domain.
The static and dynamic components can promote each other because of the complementary property between the domain-invariant and domain-specific features, in which the former ensures out-of-distribution generalization ability while the latter provides self-adaptation ability serving as a complement of the former. Therefore, to push the limits of DDG, the static and dynamic components are encouraged to be disentangled more thoroughly to obtain more comprehensive domain-invariant features and more adaptive domain-specific features, respectively.

Specifically, since the dynamic component can be further decomposed into a linear combination of static kernel templates with dynamic coefficients, the dynamic parameters are squeezed into a low-dimensional space of dynamic coefficients. Therefore, the dynamic component practically refers to the dynamic coefficients, which mainly capture the domain-specific features. To validate this argument, we visualize the dynamic coefficients using t-SNE on a DDG model trained with PACS dataset~\cite{pacs}. As shown in Figure \ref{fig:coefftsne}, the dynamic coefficients show a strong correlation with the domain labels, even without the supervision of domain labels during training. It confirms that the dynamic component, corresponding to the dynamic coefficients, can automatically capture the domain-specific information.
This evidence supports our approach of performing feature disentanglement in the low-dimensional dynamic parameter space instead of the high-dimensional feature space. By augmenting the dynamic coefficients, the static component can learn more comprehensive domain-invariant features. Alternately, the better static component can in turn facilitate the dynamic component to focus more on learning adaptive dynamic coefficients for domain-specific adaptation.
To achieve this mutually-beneficial alternation for better feature disentanglement, we propose two different \emph{Parameter Exchange (PE)} methods to perturb the dynamic component by simply exchanging dynamic coefficients in different levels.

\noindent\paragraph{\textbf{Cross-Instance Parameter Exchange}}
As shown in Figure \ref{fig:cross_sample}, cross-instance PE aims to exchange the dynamic coefficients among different instances. Practically, it is implemented by randomly shuffling the dynamic coefficients along the batch dimension, and then the shuffled dynamic coefficients are combined with the asymmetric kernel templates to assemble a newly perturbed dynamic component, which is subsequently integrated with the static component. Given an input instance $x$, the corresponding network parameters with cross-instance PE can be formulated as:
\begin{equation}
\boldsymbol{\Theta}(x) = \boldsymbol{\Theta}_s +  \sum_{m=1}^{M}{\lambda_m(\tilde{x}) \boldsymbol{\Phi}_m},
\label{eq:cross_instance}
\end{equation}
where $\{\lambda_m(\tilde{x})\}_{m=1}^M$ represents the dynamic coefficients of another instance $\tilde{x}$ in the same batch of $x$. In other words, Equation \ref{eq:cross_instance} shows that cross-instance PE replaces the dynamic coefficients of instances $x$ with those of another instance $\tilde{x}$, thereby changing the dynamic mapping from $f_{\boldsymbol{\Theta}_s+\boldsymbol{\Delta\Theta}(x)}$ to a perturbed version $f_{\boldsymbol{\Theta}_s+\boldsymbol{\Delta\Theta}(\tilde{x})}$. 
In this way, the static component $\boldsymbol{\Theta}_s$ is enforced to be invariant to the inter-instance perturbations on the dynamic component, which benefits the static component in learning more comprehensive domain-invariant features.

\noindent\paragraph{\textbf{Cross-Kernel Parameter Exchange}}
Different from cross-instance PE, cross-kernel PE aims to exchange the dynamic coefficients among kernels in an intra-instance manner. 
As shown in Figure \ref{fig:cross_kernel}, the dynamic coefficients of instance $x$ are represented as $\{\lambda_m(x)\}_{m=1}^M$, which are shuffled along the dimension of kernel templates, i.e., shuffling the sequence of
$\{\lambda_1(x), \lambda_2(x), \cdots, \lambda_M(x)\}$, to obtain a perturbed coefficients $\{\widetilde{\lambda_m}(x)\}_{m=1}^M$, which derives the perturbed dynamic component as follows:
\begin{equation}
\boldsymbol{\widetilde{\Delta\Theta}}(x) = \sum_{m=1}^{M}{\widetilde{\lambda_m}(x) \boldsymbol{\Phi}_m},
\label{eq:cross_kernel}
\end{equation}
The perturbed dynamic component is integrated with the static component to form the network parameters, and therefore, the dynamic mapping is changed from $f_{\boldsymbol{\Theta}_s+\boldsymbol{\Delta\Theta}(x)}$ to a perturbed version 
$f_{\boldsymbol{\Theta}_s+\boldsymbol{\widetilde{\Delta\Theta}}(x)}$.
In practice, each dynamic coefficient controls the dynamic magnitude of the corresponding asymmetric kernel template along the channel, height, or width dimension. By breaking and rearranging the domain-specific features, cross-kernel PE disturbs the dynamic component aggressively, which can enforce the model to learn more robust domain-invariant features.

\noindent\paragraph{\textbf{Training Objective}}
PE is implemented alternately for DDG, meaning that we update the model using gradients obtained from both non-perturbed and perturbed feed-forwards. Here, the terms ``non-perturbed'' and ``perturbed'' refer to the feed-forward operations without and with PE, respectively. The training objective of cross-instance and cross-kernel PE can be formulated as in Equation \ref{eq:ci_loss} and Equation \ref{eq:ck_loss}, respectively:

\begin{equation}
\begin{aligned}
\min\limits_{\boldsymbol{\Theta}_s, \boldsymbol{\Phi},\lambda}  \sum_{x,\tilde{x}\in \mathcal{D}_{train}}
 & \mathcal{L}_{ce}(f_{\boldsymbol{\Theta}_s+\boldsymbol{\Delta\Theta}(x)}(x), y)  \\
+ & \beta \cdot \mathcal{L}_{ce}(f_{\boldsymbol{\Theta}_s+\boldsymbol{\Delta\Theta}(\tilde{x})}(x), y),
\label{eq:ci_loss}
\end{aligned}
\end{equation}
\begin{equation}
\begin{aligned}
\min\limits_{\boldsymbol{\Theta}_s, \boldsymbol{\Phi},\lambda}  \sum_{x\in \mathcal{D}_{train}}
 & \mathcal{L}_{ce}(f_{\boldsymbol{\Theta}_s+\boldsymbol{\Delta\Theta}(x)}(x), y)  \\
+ & \beta \cdot \mathcal{L}_{ce}(f_{\boldsymbol{\Theta}_s+\boldsymbol{\widetilde{\Delta\Theta}}(x)}(x), y),
\label{eq:ck_loss}
\end{aligned}
\end{equation}
where $y$ is the class label of instance $x$, and $\mathcal{L}_{ce}$ is the standard cross-entropy loss.
$\beta$ is a hyper-parameter that controls the importance ratio of the non-perturbed feed-forward, which is set as $\beta=1$ here.

\noindent\paragraph{\textbf{Discussion}}
The collaboration of feed-forward w/ and w/o PE is very important.  On one hand, using only the perturbed feed-forward can cause the loss to fluctuate tremendously, while non-perturbed feed-forward with steady gradients can alleviate such problem and make the loss decrease more smoothly. On the other hand, the perturbed feed-forward enables the static component to capture more comprehensive domain-invariant features under perturbations from domain-specific features,  and subsequently, the non-perturbed feed-forward can facilitate the dynamic component to focus more on self-adaptive domain-specific features under the collaboration of comprehensive domain-invariant features. 

The PE method is a learning paradigm designed for the dynamic network, specifically the DDG model~\cite{sun2022ddg}. However, since there exist multiple dynamic networks with similar architectures, such as DRT~\cite{li2021dynamictransfer} and ODConv~\cite{li2022omnidimensional}, which also incorporate a dynamic component and a static component, we believe that PE can be applied to these dynamic networks as well to enhance their generalization ability. More analyses will be discussed in the experiment part.

\section{Experiments}
\subsection{Datasets and Settings} 
\noindent\paragraph{\textbf{Datasets.}} We conduct extensive experiments on multiple standard DG datasets varying from small-scale to large-scale, from small domain shift to large domain shift.
\textbf{PACS}~\cite{pacs} contains $9,991$ images with seven categories collected from four domains, i.e., \textit{Art Painting}, \textit{Cartoon}, \textit{Photo}, and \textit{Sketch}. 
\textbf{Office-Home}~\cite{venkateswara2017OfficeHome} is a more challenging dataset that contains $15,579$ images with $65$ categories, with a few images per category. All images are sampled from four domains, including: \textit{Art}, \textit{Clipart}, \textit{Product} and \textit{RealWorld}. 
\textbf{VLCS}~\cite{vlcs} consists of $10,792$ images with 5 categories, where the images are collected from \textit{Caltech} \cite{fei2004learning}, \textit{LabelMe} \cite{russell2008labelme}, \textit{PASCAL VOC 2007} \cite{everingham2010pascal}, and \textit{Sun} \cite{xiao2010sun}. All images are real photos taken by different cameras, resulting in a small domain shift in this dataset. 
\textbf{TerraIncognita}~\cite{terraincognita} contains $24,788$ images with $10$ categories of wild animals taken by camera traps at location \textit{L100}, \textit{L38}, \textit{L43}, \textit{L46}. Particularly, we follow the previous studies~\cite{gulrajani2020search_erm, kim2021selfreg, yao2022pcl} using the modified version~\cite{gulrajani2020search_erm}. 
\textbf{DomainNet}~\cite{DomainNet} is a large-scale benchmark containing nearly $600k$ images with $345$ categories. The images are distributed among six domains: \textit{Clipart}, \textit{Inforgraph}, \textit{Painting}, \textit{Quickdraw}, \textit{Real} and \textit{Sketch}. 

\noindent\paragraph{\textbf{Experimental Setup}}
We follow DDG~\cite{sun2022ddg} which adopts the leave-one-domain-out evaluation protocol, i.e., randomly selecting one domain as the target domain for evaluation and leaving the remaining domains as the source domains for training. In addition, each experiment is repeated three times with different random seeds, and the mean and variance of the experimental results are reported in this paper.

\noindent\paragraph{\textbf{Implementation Details}}
For a fair comparison, we follow similar settings in DDG~\cite{sun2022ddg} to implement our method. Our method is implemented based on a dynamic network whose backbone is ResNet-50, and the weights are initialized by the pre-trained model provided by DDG~\cite{sun2022ddg}. For most datasets, we train the dynamic network using SGD with a batch size of 64, maximum epochs of 50, an initial learning rate of 1e-3 decayed by cosine scheduler. While training with DomainNet, we keep most hyper-parameters unchanged, except that the initial learning rate is set to 2e-3 and the maximum number of epochs is set to 15, and the mini-batches are fetched with random domain sampler strategy~\cite{mixstyle}. We also follow \cite{kim2021selfreg} to adopt SWA~\cite{izmailov2018averaging} for more stable performance. Additionally, we also apply the proposed PE to other dynamic networks, including DRT~\cite{li2021dynamictransfer} and ODConv~\cite{li2022omnidimensional}, which share similar architectures with DDG. We follow the same training settings as in DDG to implement PE on these dynamic networks.

\begin{table}[t]
  \centering
  \caption{Leave-one-domain-out generalization results on PACS dataset. CI-PE and CK-PE represent cross-instance parameter exchange and cross-kernel parameter exchange, respectively. The results with \textbf{bold} and \uline{underlined} represent the best and second-best performance. The cell with gray color means the best result of PE in the corresponding dynamic backbone.
  Same representation in the following tables. 
  }
  \vspace{-10pt}
  \setlength\tabcolsep{3pt}\scalebox{0.76}{
    \begin{tabular}{l|cccc|l}
    \toprule
    Method & Art   & Cartoon &  Photo &  Sketch & Avg. \\
    \midrule
    MLDG \cite{li2018metalearningDG} & 66.23 & 66.88 & 88.00 & 58.96 & 70.01 \\
    JiGen \cite{carlucci2019JiGen} & 79.42 & 75.25 & 96.03 & 71.35 & 80.51 \\
    D-SAM \cite{d-sam} & 77.33 & 72.43 & 95.30 & 77.83 & 80.72 \\
    DANN \cite{sicilia2021dann} & 80.20 & 77.60 & 95.40 & 70.00 & 80.80 \\
    CSD \cite{piratla2020efficient_csd} & 78.90±1.10 & 75.80±1.00 & 94.10±0.20 & 76.70±1.20 & 81.40 \\
    Epi-FCR \cite{li2019epi-fcr} & 82.10 & 77.00 & 93.90 & 73.00 & 81.50 \\
    DANNCE \cite{sicilia2021dann} & 82.10 & 78.20 & 94.70 & 71.90 & 81.70 \\
    C-DANN$^+$ \cite{c-dann} & 84.60±1.80 & 75.50±0.90 & 96.80±0.30 & 73.50±0.60 & 82.60 \\
    MASF\cite{dou2019masf} & 82.89±0.16 & 80.49±0.21 & 95.01±0.10 & 72.29±0.15 & 82.67 \\
    L2A-OT \cite{zhou2020l2a_ot} & 83.30 & 78.20 & 96.20 & 73.60 & 82.80 \\
    DDAIG \cite{ddaig} & 84.20±0.30 & 78.10±0.60 & 95.30±0.40 & 74.70±0.80 & 83.10 \\
    SagNet \cite{SagNet} & 83.58 & 77.66 & 95.47 & 76.30 & 83.25 \\
    DMG \cite{chattopadhyay2020learning} & 82.57 & 78.11 & 94.49 & 78.32 & 83.37 \\
    MetaReg \cite{balaji2018metareg} & 87.20±0.13 & 79.20±0.27 & 97.60±0.31 & 70.30±0.18 & 83.60 \\
    MixStyle \cite{mixstyle} & 84.10±0.40 & 78.80±0.40 & 96.10±0.30 & 75.90±0.90 & 83.70 \\
    Mixup$^+$ \cite{mixup} & 86.10±0.50 & 78.90±0.80 & 97.60±0.10 & 75.80±1.80 & 84.60 \\
    RSC$^+$ \cite{huang2020rsc} & 85.40±0.80 & 79.70±1.80 & 97.60±0.30 & 78.20±1.20 & 85.20 \\
    ERM \cite{gulrajani2020search_erm} & 88.10±0.10 & 77.90±1.30 & 97.80±0.00 & 79.10±0.90 & 85.70 \\
    COMEN \cite{chen2022compound} & 82.60 & 81.00 & 94.60 & \textbf{84.50} & 85.70 \\
    SelfReg \cite{kim2021selfreg} & 85.90±0.60 & 81.90±0.40 & 96.80±0.10 & 81.40±0.60 & 86.50 \\
    SWAD \cite{cha2021swad} & \textbf{89.30±0.20} & 83.40±0.60 & 97.30±0.30 & 82.50±0.50 & 88.10 \\
    \midrule
    DDG \cite{sun2022ddg} & 85.92±0.17 & 79.68±0.42 & 96.65±0.15 & 82.62±0.27 & 86.21 \\
    DDG w/ CI-PE (Ours) & 86.48±0.63 & 81.48±0.58 & 97.72±0.08 & 82.74±0.15 & 87.10$\uparrow$ \\
    DDG w/ CK-PE (Ours) & 87.07±0.55 & 82.89±0.30 & 97.83±0.19 & 82.03±0.19 & \cellcolor{gray!40}87.46$\uparrow$ \\
    \midrule
    DRT \cite{li2021dynamictransfer} & 85.13±0.59 & 77.07±0.46 & 98.16±0.06 & 80.08±0.25 & 85.11 \\
    DRT w/ CI-PE (Ours) & 86.42±0.25 & 79.80±0.39 & 98.06±0.08 & 81.09±0.13 & 86.34$\uparrow$ \\
     DRT w/ CK-PE (Ours) & 87.89±0.67 & 80.13±0.56 & 97.88±0.12 & 80.89±0.11 & \cellcolor{gray!40}86.70$\uparrow$ \\
    \midrule
    ODConv \cite{li2022omnidimensional} & 86.51±0.17 & 79.82±0.56 & 97.98±0.08 & 80.68±0.77 & 86.25 \\
    ODConv w/ CI-PE (Ours) & 87.37±0.31 & \uline{83.64±0.58} & \uline{98.20±0.18} & \uline{83.79±0.89} & \uline{88.25}$\uparrow$ \\
     ODConv w/ CK-PE (Ours) & \uline{88.71±0.47} & \textbf{84.96±0.59} & \textbf{98.24±0.22} & 82.04±0.29 & \cellcolor{gray!40}\textbf{88.49}$\uparrow$ \\
    \bottomrule
    \end{tabular}%
  }
  \label{tab:pacs}%
  \vskip -0.10in
\end{table}%
\begin{table}[t]
  \centering
    \caption{Leave-one-domain-out results on Office-Home.}\label{tab:office}%
    \vspace{-10pt}
  \setlength\tabcolsep{3pt}\scalebox{0.76}{
    \begin{tabular}{l|cccc|l}
    \toprule
    Method & Art   &  Clipart & Product & RealWorld & Avg. \\
    \midrule
    D-SAM \cite{d-sam} & 58.03 & 44.37 & 69.22 & 71.45 & 60.77 \\
    JiGen \cite{carlucci2019JiGen} & 53.04 & 47.51 & 71.47 & 72.79 & 61.20 \\
    SagNet \cite{SagNet} & 60.20 & 45.38 & 70.42 & 73.38 & 62.34 \\
    MixStyle \cite{mixstyle} & 58.70±0.30 & 53.40±0.20 & 74.20±0.10 & 75.90±0.10 & 65.50 \\
    RSC$^+$ \cite{huang2020rsc} & 60.70±1.40 & 51.40±0.30 & 74.80±1.10 & 75.10±1.30 & 65.50 \\
    DDAIG \cite{ddaig} & 59.20±0.10 & 52.30±0.30 & 74.60±0.30 & 76.00±0.10 & 65.50 \\
    L2A-OT \cite{zhou2020l2a_ot} & 60.60 & 50.10 & 74.80 & 77.00 & 65.60 \\
    DANN \cite{sicilia2021dann} & 61.60 & 48.90 & 75.80 & 76.20 & 65.60 \\
    C-DANN$^+$ \cite{c-dann} & 61.50±1.40 & 50.40±2.40 & 74.40±0.90 & 76.60±0.80 & 65.80 \\
    DANNCE \cite{sicilia2021dann} & 61.60 & 50.20 & 75.60 & 75.90 & 65.80 \\
    COMEN \cite{chen2022compound} & 57.60 & 55.80 & 75.50 & 76.90 & 66.50 \\
    MLDG$^+$ \cite{li2018metalearningDG} & 61.50±0.90 & 53.20±0.60 & 75.00±1.20 & 77.50±0.40 & 66.80 \\
    ERM \cite{gulrajani2020search_erm} & 62.70±1.10 & 53.40±0.60 & 76.50±0.40 & 77.30±0.30 & 67.50 \\
    Mixup$^+$ \cite{mixup} & 62.40±0.80 & 54.80±0.60 & 76.90±0.30 & 78.30±0.20 & 68.10 \\
    SelfReg \cite{kim2021selfreg} & 64.90±0.80 & 55.40±0.60 & 78.40±0.20 & 78.80±0.10 & 69.40 \\
    SWAD \cite{cha2021swad} & 66.10±0.40 & 57.70±0.40 & 78.40±0.10 & 80.20±0.20 & 70.60 \\
    \midrule
    DDG \cite{sun2022ddg} & 69.39±0.07 & 57.15±0.17 & 80.20±0.19 & 81.79±0.07 & 72.13 \\
    DDG w/ CI-PE (Ours) & 68.61±0.15 & 58.93±0.34 & \uline{80.82±0.13} & 81.76±0.12 & 72.53$\uparrow$ \\
    DDG w/ CK-PE (Ours) & 68.47±0.23 & 59.85±0.22 & 80.57±0.09 & 81.31±0.17 & \cellcolor{gray!40}72.55$\uparrow$ \\
    \midrule
    DRT \cite{li2021dynamictransfer} & 68.16±0.17 & 58.72±0.18 & 79.67±0.20 & 81.02±0.15 & 71.89 \\
    DRT w/ CI-PE (Ours) & 69.24±0.25 & 59.28±0.15 & 80.10±0.13 & 81.15±0.08 & \cellcolor{gray!40}72.44$\uparrow$ \\
    DRT w/ CK-PE (Ours) & 68.27±0.25 & 59.83±0.36 & 80.26±0.18 & 81.34±0.12 & 72.43$\uparrow$ \\
    \midrule
    ODConv \cite{li2022omnidimensional} & 69.15±0.41 & 58.89±0.35 & 80.33±0.15 & \uline{82.72±0.11} & 72.77 \\
    ODConv w/ CI-PE (Ours) & \uline{70.58±0.33} & \textbf{60.95±0.29} & \textbf{80.93±0.35} & \textbf{82.96±0.23} & \cellcolor{gray!40}\textbf{73.86}$\uparrow$ \\
    ODConv w/ CK-PE (Ours) & \textbf{70.91±0.18} & \uline{60.07±0.19} & 80.37±0.17 & 82.38±0.22 & \uline{73.43}$\uparrow$ \\
    \bottomrule
    \end{tabular}%
  }
\end{table}%
\vspace{-0.1cm}
\begin{table}[t]
  \centering
  \caption{Leave-one-domain-out results on VLCS.}
   \vspace{-10pt}
  \label{tab:vlcs}%
  \setlength\tabcolsep{3pt}\scalebox{0.76}{
    \begin{tabular}{l|cccc|l}
    \toprule
    Method & Caltech & Labelme & Pascal & Sun   & Avg. \\
    \midrule
    Epi-FCR \cite{li2019epi-fcr} & 94.10 & 64.30 & 67.10 & 65.90 & 72.90 \\
    JiGen \cite{carlucci2019JiGen} & 96.93 & 60.90 & 70.62 & 64.30 & 73.19 \\
    MASF \cite{dou2019masf} & 94.78±0.16 & 64.90±0.08 & 69.14±0.19 & 67.64±0.12 & 74.11 \\
    COMEN \cite{chen2022compound} & 97.00 & 62.60 & 72.80 & 67.60 & 75.00 \\
    RSC$^+$ \cite{huang2020rsc} & 97.90±0.10 & 62.50±0.70 & 75.60±0.80 & 72.30±1.20 & 77.10 \\
    Mixup$^+$ \cite{mixup} & 98.30±0.60 & 64.80±1.00 & 74.30±0.80 & 72.10±0.50 & 77.40 \\
    MLDG$^+$ \cite{li2018metalearningDG} & 97.40±0.20 & \textbf{65.20±0.70} & 75.30±1.00 & 71.00±1.40 & 77.20 \\
    ERM \cite{gulrajani2020search_erm} & 97.70±0.40 & 64.30±0.90 & 74.60±1.30 & \uline{73.40±0.50} & 77.50 \\
    C-DANN$^+$ \cite{c-dann} & 97.10±0.30 & \uline{65.10±1.20} & 77.10±1.50 & 70.70±0.80 & 77.50 \\
    SelfReg \cite{kim2021selfreg} & 97.40±0.40 & 63.50±0.30 & 76.70±0.70 & 72.60±0.10 & 77.50 \\
    SagNet$^+$ \cite{SagNet} & 97.90±0.40 & 64.50±0.50 & 77.50±0.50 & 71.40±1.30 & 77.80 \\
    DANN$^+$ \cite{sicilia2021dann} & 99.00±0.30 & \uline{65.10±1.40} & 75.60±0.80 & 72.30±1.20 & 78.60 \\
    SWAD \cite{cha2021swad} & 98.80±0.10 & 63.30±0.30 & 79.20±0.60 & \textbf{75.30±0.50} & \textbf{79.10} \\
    \midrule
    DDG \cite{sun2022ddg} & 97.55±0.24 & 62.96±0.51 & 74.63±0.44 & 71.15±0.56 & 76.57 \\
    DDG w/ CI-PE (Ours) & \textbf{99.53±0.00} & 62.91±0.84 & 76.68±0.31 & 70.78±0.31 & 77.48$\uparrow$ \\
    DDG w/ CK-PE (Ours) & 98.54±0.18 & 62.81±0.47 & 77.83±0.20 &  71.80±0.38 & \cellcolor{gray!40}77.74$\uparrow$ \\
    \midrule
    DRT \cite{li2021dynamictransfer} & 97.92±0.28 & 62.89±0.55 & 76.98±0.30 & 69.08±0.43 & 76.72 \\
    DRT w/ CI-PE (Ours) & 98.72±0.32 & 63.21±0.64 & 76.50±0.42 & 69.73±0.39 & 77.04$\uparrow$ \\
    DRT w/ CK-PE (Ours) & 99.15±0.24 & 63.39±0.41 & 76.90±0.30 & 68.89±0.26 & \cellcolor{gray!40}77.08$\uparrow$ \\
    \midrule
    ODConv \cite{li2022omnidimensional} & \uline{99.48±0.10} & 62.54±0.21 & 77.49±0.43 & 71.98±0.57 & 77.87 \\
    ODConv w/ CI-PE (Ours) & \uline{99.48±0.10} & 63.91±0.15 & \uline{79.43±0.27} & 71.78±0.25 & 78.65$\uparrow$ \\
    ODConv w/ CK-PE (Ours) & 99.29±0.26 & 64.14±0.42 & \textbf{79.53±0.48} & 71.94±0.19 & \cellcolor{gray!40}\uline{78.72}$\uparrow$ \\
    \bottomrule
    \end{tabular}%
  }
  \vspace{-0.4cm}
\end{table}%
\begin{table}[!t]
  \centering
   \caption{Leave-one-domain-out results on TerraIncognita.}
   \vspace{-10pt}
  \label{tab:terra}%
  \setlength\tabcolsep{3pt}\scalebox{0.76}{
    \begin{tabular}{l|cccc|l}
    \toprule
    Method & L100  & L38   & L43   & L46   & Avg. \\
    \midrule
    C-DANN$^+$ \cite{c-dann} & 47.00±1.90 & 41.30±4.80 & 54.90±1.70 & 39.80±2.30 & 45.80 \\
    ERM \cite{gulrajani2020search_erm} & 49.80±4.40 & 42.10±1.40 & 56.90±1.80 & 35.70±3.90 & 46.10 \\
    RSC$^+$ \cite{huang2020rsc} & 50.20±2.20 & 39.20±1.40 & 56.30±1.40 & 40.80±0.60 & 46.60 \\
    DANN$^+$ \cite{sicilia2021dann} & 51.10±3.50 & 40.60±0.60 & 57.40±0.50 & 37.70±1.80 & 46.70 \\
    MLDG$^+$ \cite{li2018metalearningDG} & 54.20±3.00 & 44.30±1.10 & 55.60±0.30 & 36.90±2.20 & 47.70 \\
    Mixup$^+$ \cite{mixup} & 59.60±2.00 & 42.20±1.40 & 55.90±0.80 & 33.90±1.40 & 47.90 \\
    SagNet$^+$ \cite{SagNet} & 53.00±2.90 & 43.00±2.50 & 57.90±0.60 & 40.40±1.30 & 48.60 \\
    SWAD \cite{cha2021swad} & 55.40±0.00 & 44.90±1.10 & 59.70±0.40 & 39.90±0.20 & 50.00 \\
    SelfReg \cite{kim2021selfreg} & 56.80±0.90 & 44.70±0.60 & 59.60±0.30 & 42.90±0.80 & 51.00 \\
    \midrule
    DDG\cite{sun2022ddg} & \textbf{65.47±0.35} & 42.55±1.25 & 53.90±0.55 & 35.84±0.67 & 49.44 \\
    DDG w/ CI-PE (Ours) & 56.24±0.95 & 48.82±0.40 & 59.71±0.47 & 44.65±0.33 & \cellcolor{gray!40}52.35$\uparrow$ \\
    DDG w/ CK-PE (Ours) & 58.78±0.30 & 45.92±0.93 & 60.17±0.14 & 43.94±0.51 & 52.20$\uparrow$ \\
    \midrule
    DRT \cite{li2021dynamictransfer} & 53.02±1.04 & \textbf{50.12±0.68} & 55.34±0.33 & 41.01±0.22 & 49.87 \\
    DRT w/ CI-PE (Ours) & 52.74±1.61 & \uline{49.22±0.42} & 59.55±0.18 & 43.26±0.41 & \cellcolor{gray!40}51.19$\uparrow$ \\
    DRT w/ CK-PE (Ours) & 54.12±1.15 & 48.59±0.11 & 58.48±0.27 & 42.15±0.44 & 50.84$\uparrow$ \\
    \midrule
    ODConv \cite{li2022omnidimensional} & 48.77±1.84 & 40.44±0.56 & 58.68±0.19 & 44.22±0.51 & 48.03 \\
    ODConv w/ CI-PE (Ours) & 58.71±1.05 & 45.88±1.03 & \uline{60.75±0.43} & \textbf{49.01±0.40} & \cellcolor{gray!40}\textbf{53.59}$\uparrow$ \\
    ODConv w/ CK-PE (Ours) & \uline{59.88±0.78} & 42.21±1.37 & \textbf{60.93±0.23} & \uline{47.98±0.29} & \uline{52.75}$\uparrow$ \\
    \bottomrule
    \end{tabular}%
  }
\end{table}%

\begin{table*}[t]
  \centering
    \caption{Leave-one-domain-out generalization results on DomainNet. Our method achieves the state-of-the-art performance.}
   \vspace{-10pt}
  \label{tab:domainnet}%
  \setlength\tabcolsep{17.5pt}\scalebox{0.76}{
    \begin{tabular}{l|cccccc|l}
    \toprule
    Method & Clipart &  Infograph &  Painting & Quickdraw & Real  &  Sketch & Avg. \\
    \midrule
    C-DANN$^+$ \cite{c-dann} & 54.60±0.40 & 17.30±0.10 & 43.70±0.90 & 12.10±0.70 & 56.20±0.40 & 45.90±0.50 & 38.30 \\
    DANN$^+$ \cite{sicilia2021dann} & 53.10±0.20 & 18.30±0.10 & 44.20±0.70 & 11.80±0.10 & 55.50±0.40 & 46.80±0.60 & 38.30 \\
    RSC$^+$ \cite{huang2020rsc} & 55.00±1.20 & 18.30±0.50 & 44.40±0.60 & 12.20±0.20 & 55.70±0.70 & 47.80±0.90 & 38.90 \\
    Mixup$^+$ \cite{mixup} & 55.70±0.30 & 18.50±0.50 & 44.30±0.50 & 12.50±0.40 & 55.80±0.30 & 48.20±0.50 & 39.20 \\
    SagNet$^+$ \cite{SagNet} & 57.70±0.30 & 19.00±0.20 & 45.30±0.30 & 12.70±0.50 & 58.10±0.50 & 48.80±0.20 & 40.30 \\
    ERM \cite{gulrajani2020search_erm} & 58.10±0.30 & 18.80±0.30 & 46.70±0.30 & 12.20±0.40 & 59.60±0.10 & 49.80±0.40 & 40.90 \\
    MLDG$^+$ \cite{li2018metalearningDG} & 59.10±0.20 & 19.10±0.30 & 45.80±0.70 & 13.40±0.30 & 59.60±0.20 & 50.20±0.40 & 41.20 \\
    MetaReg \cite{balaji2018metareg} & 59.77 & 25.58 & 50.19 & 11.52 & 64.56 & 50.09 & 43.62 \\
    DMG \cite{chattopadhyay2020learning} & 65.24 & 22.15 & 50.03 & 15.68 & 59.63 & 49.02 & 43.63 \\
    SelfReg \cite{kim2021selfreg} & 62.40±0.10 & 22.60±0.10 & 51.80±0.10 & 14.30±0.10 & 62.50±0.20 & 53.80±0.30 & 44.60 \\
    SWAD \cite{cha2021swad} & 66.00±0.10 & 22.40±0.20 & 53.50±0.10 & 16.10±0.20 & 65.80±0.40 & \uline{55.50±0.30} & 46.50 \\
    \midrule
    DDG \cite{sun2022ddg} & 68.07±0.08 & 25.02±0.18 & 53.54±0.37 & 13.74±0.10 & \textbf{67.39±0.02} & 53.59±0.06 & 46.89 \\
    DDG w/ CI-PE (Ours) & 68.93±0.24 & 25.61±0.05 & 55.74±0.14 & 17.57±0.18 & 66.61±0.17 & \textbf{55.83±0.21} & \cellcolor{gray!40}48.38$\uparrow$ \\
    DDG w/ CK-PE (Ours) & 69.25±0.13 & 25.41±0.18 & 55.90±0.18 & 16.63±0.24 & 66.77±0.09 & 55.27±0.19 & 48.20$\uparrow$ \\
    \midrule
    DRT \cite{li2021dynamictransfer} & 67.69±0.17 & 24.65±0.13 & 53.10±0.57 & 14.56±0.29 & 65.95±0.19 & 53.00±0.19 & 46.49 \\
    DRT w/ CI-PE (Ours) & 69.45±0.26 & 25.54±0.23 & 55.74±0.15 & 16.69±0.14 & 66.67±0.11 & 55.19±0.20 & \cellcolor{gray!40}48.21$\uparrow$ \\
    DRT w/ CK-PE (Ours) & 69.39±0.24 & 25.92±0.10 & 56.05±0.11 & 16.78±0.17 & 66.82±0.19 & 54.11±0.37 & 48.18$\uparrow$ \\
    \midrule
    ODConv \cite{li2022omnidimensional} & 69.04±0.43 & 25.60±0.16 & 54.46±0.23 & 15.03±0.34 & \uline{67.27±0.09} & 53.04±0.47 & 47.41 \\
    ODConv w/ CI-PE (Ours) & \uline{69.74±0.22} & \textbf{27.81±0.14} & \uline{57.04±0.26} & \textbf{18.09±0.25} & 67.03±0.19 & 54.55±0.36 & \cellcolor{gray!40}\textbf{49.04}$\uparrow$ \\
    ODConv w/ CK-PE (Ours) & \textbf{69.97±0.22} & \uline{27.30±0.20} & \textbf{57.34±0.23} & \uline{17.94±0.19} & 66.76±0.28 & 53.69±0.47 & \uline{48.83}$\uparrow$ \\
    \bottomrule
    \end{tabular}%
  }
  \vspace{-0.3cm}
\end{table*}%

\subsection{Comparisons with the State-of-the-Arts}\label{sec:sota}
We compare our method with the state-of-the-art DG methods on the PACS, Office-Home, VLCS, TerraIncognita, and DomainNet datasets. The proposed parameter exchange methods, i.e., cross-instance PE (named \textbf{CI-PE} for short) and cross-kernel PE (named \textbf{CK-PE} for short), are implemented on the basis of different dynamic networks, i.e., DDG~\cite{sun2022ddg}, DRT~\cite{li2021dynamictransfer} and ODConv~\cite{li2022omnidimensional}.
The comparison results on different DG datasets are presented in Table \ref{tab:pacs}, Table \ref{tab:office}, Table \ref{tab:vlcs}, Table \ref{tab:terra} and Table \ref{tab:domainnet}, respectively. 
Some results marked by $^+$ are re-implemented and reported by~\protect\cite{gulrajani2020search_erm}. 
From these tables, we can discover and summarize that: 
\vspace{-\topsep}
\begin{itemize}[leftmargin=12pt, topsep=6pt, itemsep=3pt]
    \item The two proposed PE methods, cross-instance PE and cross-kernel PE, exhibit stable performance improvement across all DG datasets compared to strong baseline models such as DDG, DRT, and ODConv. This strongly indicates the effectiveness of PE in enhancing dynamic domain generalization and its potential as a plug-and-play method for various dynamic networks to improve their generalization ability. 
    \item Our method achieves excellent performances on all datasets, especially obtaining the state-of-the-art results on PACS, Office-Home, TerraIncognita and DomainNet datasets, which shows the significant effectiveness of our method in DG task. Besides, our method involves less supervision signals that do not need to use domain labels, which is different from most DG approaches.
    \item Although the two proposed PE methods can consistently enhance DDG, they have differences on the way to acquire newly-stylized domain. While cross-instance PE employs domain information from other instances, cross-kernel PE  synthesizes a new implicit domain by rearranging and exchanging of the dynamic coefficients that holds valuable domain-specific information. Such differences are also confirmed by the experimental results. For large- and medium-scale datasets such as Office-Home, TerraIncognita, and DomainNet, the performance of cross-instance PE outperforms that of cross-kernel PE. On the other hand, for small-scale datasets like PACS and VLCS, cross-kernel PE is found to be more effective than cross-instance PE. The reason for these observations is likely due to the fact that larger datasets offer more diversity for inter-instance perturbations through cross-instance PE, resulting in better performance. Conversely, smaller datasets have limited diversity for inter-instance exchanging, making cross-kernel PE a more suitable choice for small-scale datasets.
\end{itemize}
\vspace{-\topsep}

\subsection{Ablation Studies}
\begin{table}[!t]
  \centering
    \caption{Ablation studies on CI-PE, CK-PE, and SWA. }
   \vspace{-10pt}
  \label{tab:ablation_pacs_vlcs}
  \setlength\tabcolsep{16pt}\scalebox{0.75}{
    \begin{tabular}{ccc|cc|c}
    \toprule
    \multicolumn{3}{c|}{Component} & \multicolumn{2}{c|}{Dataset} & \multirow{2}{*}{Avg.} \\
     CI-PE & CK-PE & SWA & PACS  & VLCS\\
    \midrule
     \textbf{-} & \textbf{-} & \textbf{-} & 86.21 & 76.57 & 81.39 \\
     \textbf{-} & \textbf{-} & $\checkmark$ & 85.04 & 75.70 & 80.37 \\
    \midrule
    $\checkmark$ & \textbf{-} & \textbf{-} & 86.96 & 76.97 & 81.97 \\
    $\checkmark$ & \textbf{-} & $\checkmark$ & 87.10 & 77.48 & 82.29 \\ 
    \midrule
     \textbf{-} & $\checkmark$ & \textbf{-} & 86.78 & 77.24 & 82.01 \\
    \textbf{-} & $\checkmark$ & $\checkmark$ & \textbf{87.46} & \textbf{77.74} & \textbf{82.60} \\
    \bottomrule
    \end{tabular}%
    }
 \vspace{-0.2cm}
\end{table}%
\begin{table}[t]
  \centering
    \caption{Ablation studies of different parameter perturbation manners,  i.e., parameter exchange and parameter mix, on PACS dataset. Note that these implementation are only performed in the way of cross-instance.}
   \vspace{-10pt}
  \label{tab:ablation_combinetype}%
  \setlength\tabcolsep{4pt}\scalebox{0.75}{
    \begin{tabular}{c|cccc|c}
    \toprule
    \multicolumn{1}{l|}{Updating Manner} & Art & Cartoon & Photo & Sketch & \multicolumn{1}{l}{Avg.} \\
    \midrule
    Baseline \cite{sun2022ddg} & 85.92±0.17 & 79.68±0.42 & 96.65±0.15 & 82.62±0.27 & 86.21 \\
    Parameter Mix   & 85.66±0.86 & 80.86±0.52 & 97.29±0.13 & \textbf{82.93±0.36} & 86.69 \\
    Parameter Exchange & \textbf{86.48±0.63} & \textbf{81.48±0.58} & \textbf{97.72±0.08} & 82.74±0.15 & \textbf{87.10} \\
    \bottomrule
    \end{tabular}%
  }
  \vspace{-0.2cm}
\end{table}%
\begin{table}[htbp]
  \centering
\caption{Ablation studies of different exchange strategies in terms of cross-instance PE on PACS dataset.}
   \vspace{-10pt}
  \label{tab:exc_stgy}%
    \setlength\tabcolsep{5pt}\scalebox{0.75}{
    \begin{tabular}{c|cccc|c}
    \toprule
    Exchange Strategy & Art   & Cartoon & Photo & Sketch & Avg. \\
    \midrule
    w/ same class & 85.84±0.36 & 78.75±0.71 & 97.25±0.15 & 81.28±0.64 & 85.78 \\
    w/ different class & 85.24±0.48 & \textbf{82.48±0.71} & 97.60±0.10 & 82.67±0.46 & 87.00 \\
    w/ same domain & 85.16±0.29 & 81.04±0.64 & 97.39±0.16 & 80.41±0.33 & 86.00 \\
    w/ different domain & 85.23±0.54 & 81.67±0.55 & 97.02±0.21 & 81.30±0.65 & 86.31 \\
    w/ random shuffle & \textbf{86.48±0.63} & 81.48±0.58 & \textbf{97.72±0.08} & \textbf{82.74±0.15} & \textbf{87.10} \\
    \bottomrule
    \end{tabular}%
    }
    \vspace{-0.3cm}
\end{table}%

We conduct ablation studies on PACS and VLCS datasets to investigate the effectiveness and properties of the proposed parameter exchange methods on the basis of DDG~\cite{sun2022ddg}.

\noindent\paragraph{\textbf{Exploring Different Components in Parameter Exchange}}
To further investigate the effectiveness of our method more thoroughly, we implement cross-instance PE and cross-kernel PE with and without SWA~\cite{izmailov2018averaging}. For a fair comparison, we also implement a new variant of the baseline that trains DDG with SWA. The comparison results are shown in Table \ref{tab:ablation_pacs_vlcs}.  Notably, we found that SWA does not provide any improvement when used alone with DDG on PACS and VLCS datasets. However, when integrated with our PE method, SWA is shown to further enhance the already strong performance of cross-instance PE or cross-kernel PE, indicating the effectiveness of combining PE with SWA. Our findings demonstrate that the success of our approach is not solely due to the use of SWA, but rather from the synergistic combination of PE and SWA.

\noindent\paragraph{\textbf{Exploring the Effectiveness of Different Parameter Perturbation Manners}}
We investigate the effectiveness of different parameter perturbation manners for PE. 
Different from cross-instance PE, we also study another parameter perturbation manner, namely \emph{parameter mixing}, in which the dynamic parameters (i.e., dynamic coefficients) are exchanged and mixed across different instances in the manner of a random probabilistic convex combination:
$\lambda^{'}_m(x) = \eta \lambda_m(x) + (1-\eta) \lambda_m(\tilde{x})$,
where $\eta$ is random value.
Table \ref{tab:ablation_combinetype} shows the comparison results among parameter exchanging and parameter mixing on PACS. It turns out that the parameter exchange method slightly outperforms the parameter mixing. When compared to the subtle perturbation of parameter mixing, the parameter exchange method is capable of introducing a greater degree of perturbation on the dynamic component of the model, resulting in larger improvements in the performance of DDG. 
For this reason, we have selected parameter exchange as our primary method of perturbation.
\noindent\paragraph{\textbf{Exploring the Effectiveness of Different Exchange Strategies}}
In addition to the random exchange strategy in cross-instance PE, we also investigate other alternative exchange strategies such as exchanging dynamic coefficients between instances with the same class (wSC), different class (wDC), same domain (wSD), different domain (wDD), or through random shuffling (wRand). The results of comparing these different exchange strategies are presented in Table \ref{tab:exc_stgy}. The findings indicate that the wRand and wDC strategies exhibit comparable performances and outperform the other strategies, since the wDC and wRand strategies have greater intra-batch perturbations than other strategies. These results suggest that more thorough instance exchange can lead to more diverse perturbations, thereby improving generalization ability. Therefore, we adopt the wRand strategy as the default exchange strategy in our experiments.

\subsection{In-Depth Analysis}

\setlength{\fboxsep}{0pt}
\begin{figure}[t]
     \centering
     \setlength{\fboxrule}{0.2pt}
     \begin{subfigure}[b]{0.315\columnwidth}
         \centering
         \fbox{\includegraphics[width=\textwidth, height=1.78cm]{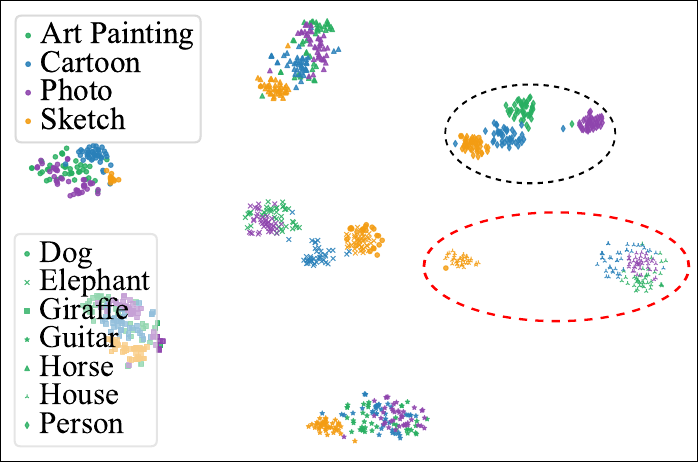}}
        \vspace{-0.5cm}
         \caption{DDG (Baseline)}
         \label{fig:static_ddg}
     \end{subfigure}
     \hfill
     \begin{subfigure}[b]{0.315\columnwidth}
         \centering
         \fbox{\includegraphics[width=\textwidth, height=1.78cm]{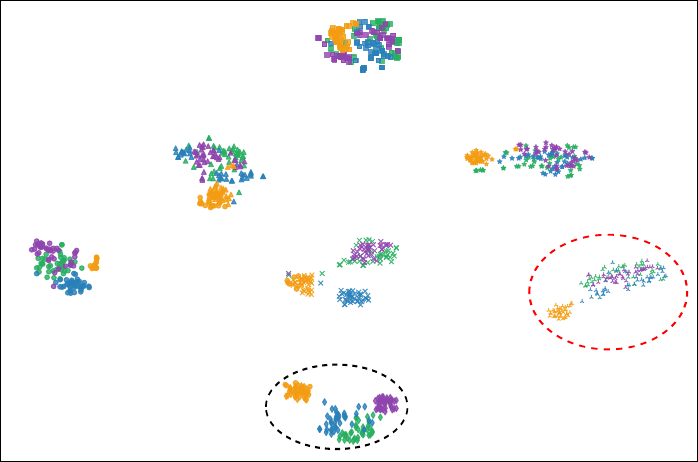}}
        \vspace{-0.5cm}
         \caption{CI-PE (Ours)}
         \label{fig:static_cipe}
     \end{subfigure}
     \hfill
     \begin{subfigure}[b]{0.315\columnwidth}
         \centering
         \fbox{\includegraphics[width=\textwidth, height=1.78cm]{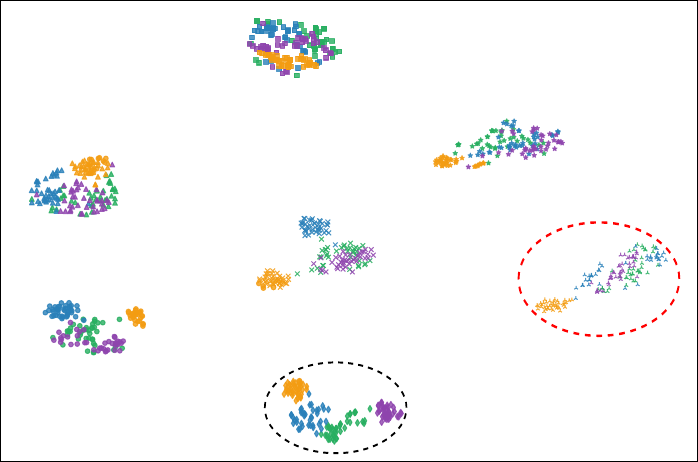}}
        \vspace{-0.5cm}
         \caption{CK-PE (Ours)}
         \label{fig:static_ckpe}
     \end{subfigure}\\
     \setlength{\fboxrule}{0.3pt}
    \begin{subfigure}[b]{0.315\columnwidth}
         \centering
         \fbox{\includegraphics[width=\textwidth, height=1.78cm]{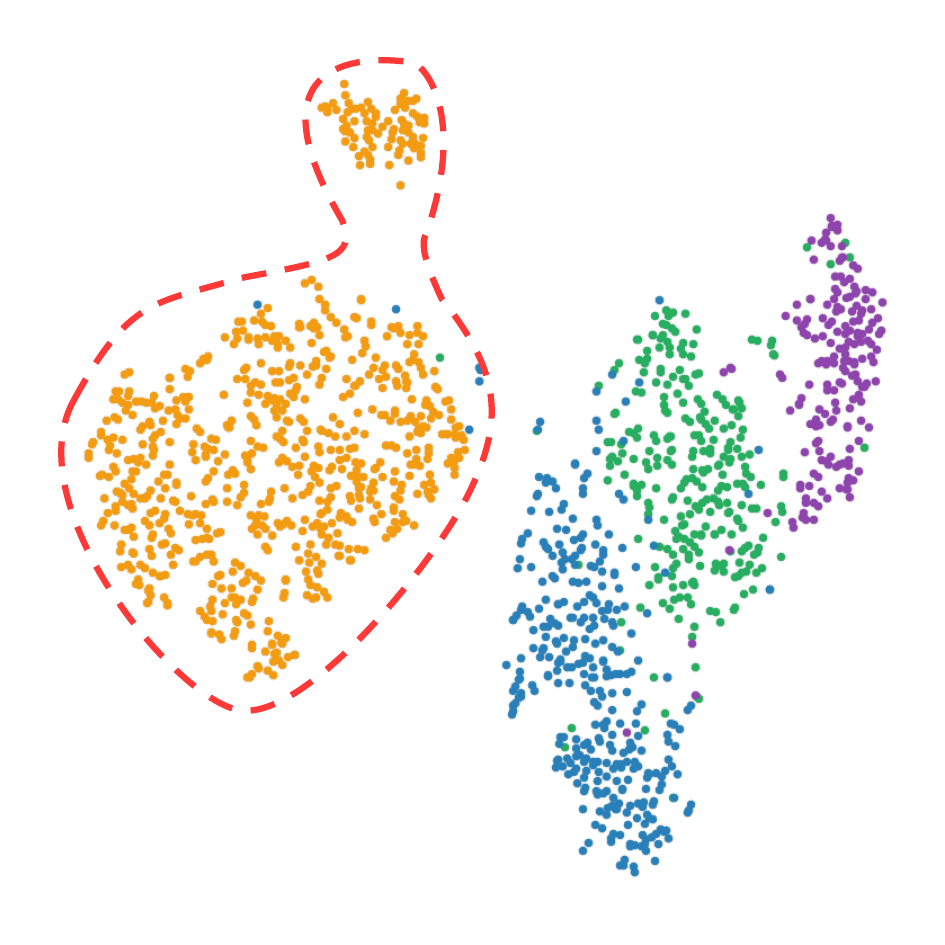}}
        \vspace{-0.5cm}
         \caption{DDG (Baseline)}
         \label{fig:dynamic_ddg}
    \end{subfigure}
     \hfill
    \begin{subfigure}[b]{0.315\columnwidth}
         \centering
         \fbox{\includegraphics[width=\textwidth, height=1.78cm]{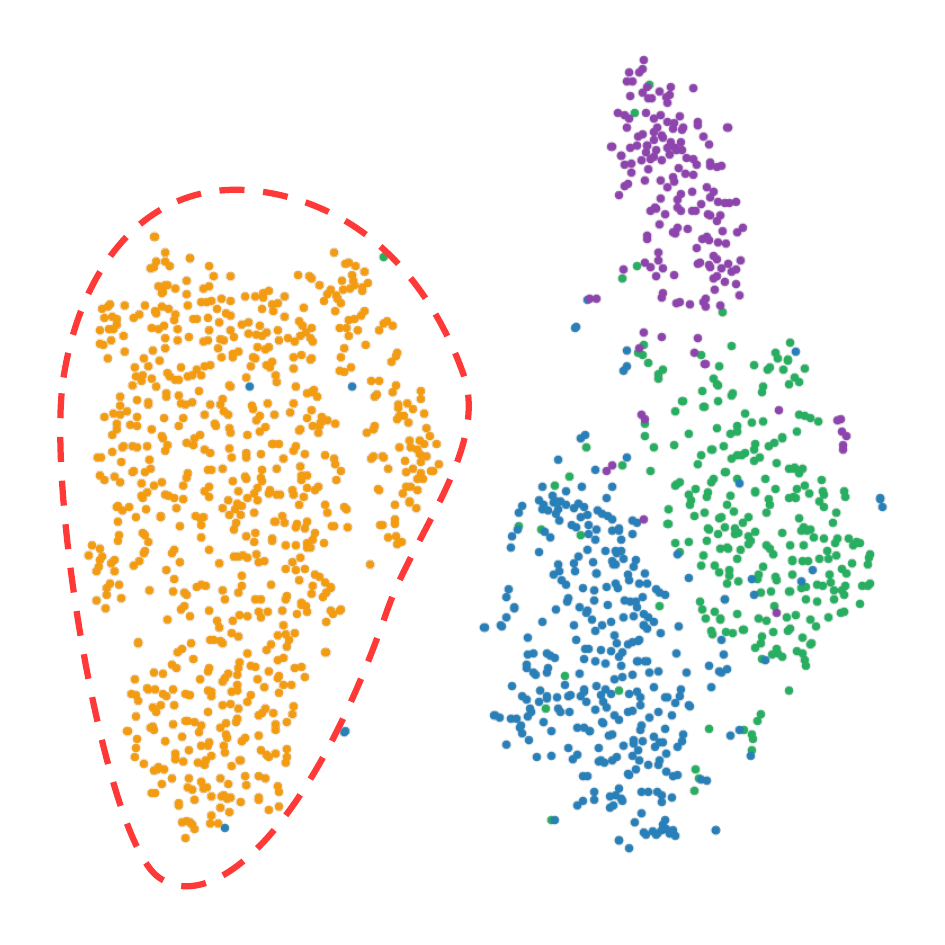}}
        \vspace{-0.5cm}
         \caption{CI-PE (Ours)}
         \label{fig:dynamic_cipe}
    \end{subfigure}
     \hfill
    \begin{subfigure}[b]{0.315\columnwidth}
         \centering
         \fbox{\includegraphics[width=\textwidth, height=1.78cm]{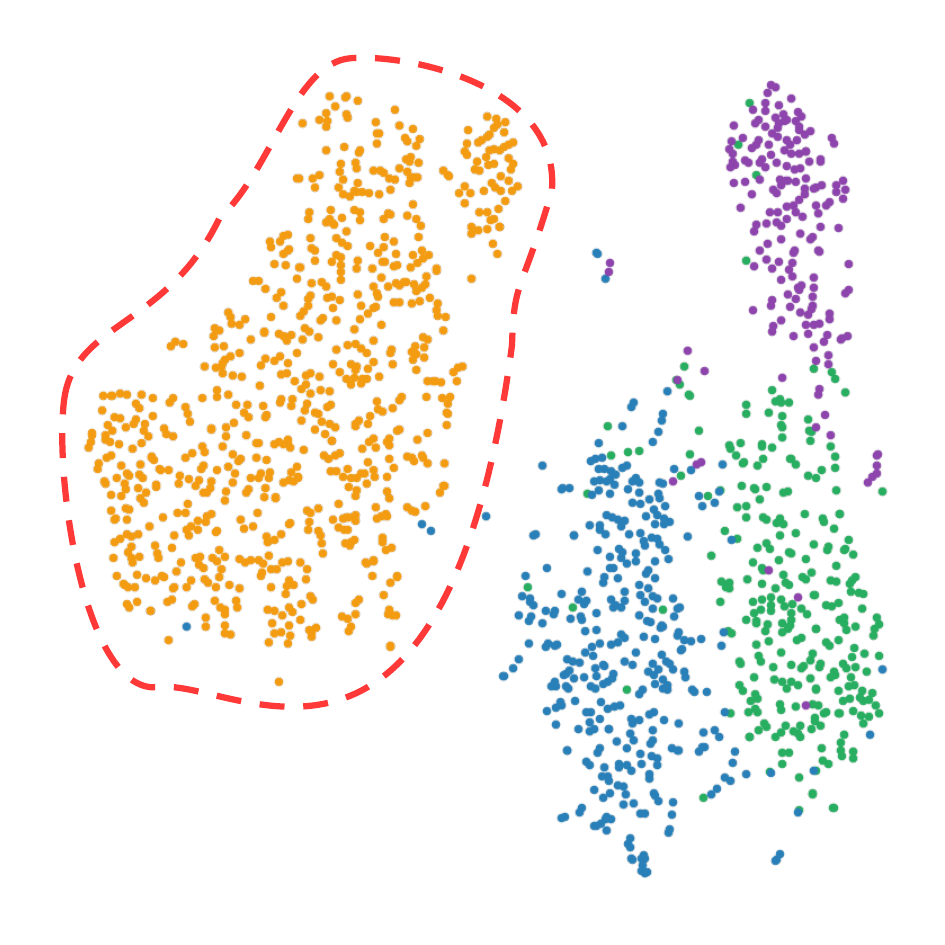}}
        \vspace{-0.5cm}
         \caption{CK-PE (Ours)}
         \label{fig:dynamic_ckpe}
    \end{subfigure}\\ 
    \vspace{-0.3cm}
        \caption{t-SNE visualization of the \uline{static and dynamic components} of different models trained with DDG, CI-PE, or CK-PE methods on the Art, Cartoon, and Photo domains of the PACS dataset, and evaluated on the Sketch domain. (a)-(c) displays the features provided by the static component of the last block in each model. (d)-(f) visualizes the dynamic coefficients of the dynamic component. The colors in the plots represent different domains. Best viewed in colors.}
        \label{fig:tsne_features}
\end{figure}
\begin{figure}[t]
     \centering
     \begin{subfigure}[b]{0.45\columnwidth}
         \centering
         \includegraphics[width=\textwidth]{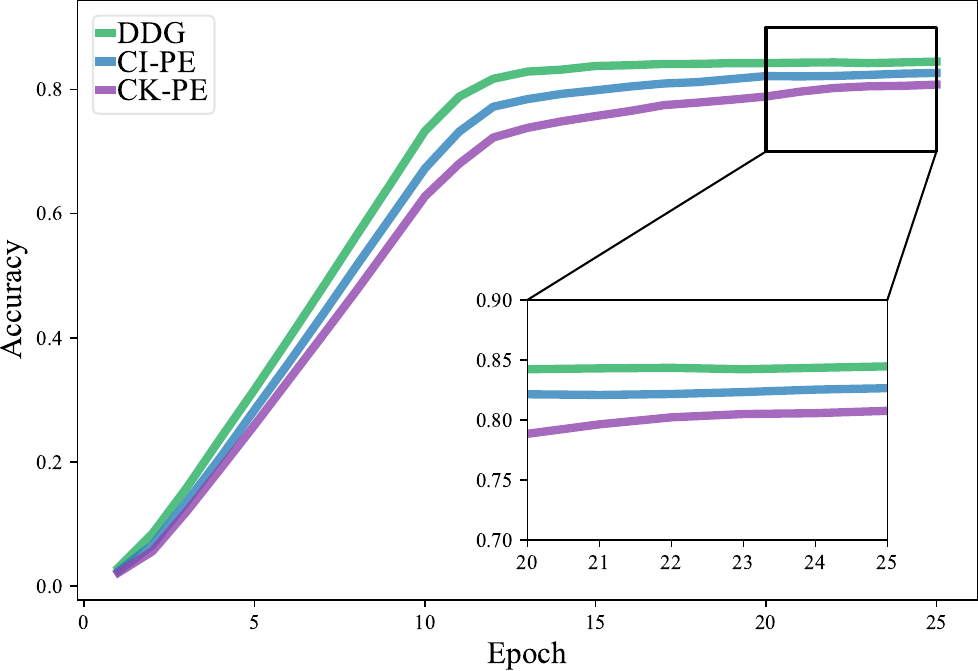}
        \vspace{-0.5cm}
         \caption{Domain classification accuracy of static features}
         \label{fig:domain_accuracy_static}
     \end{subfigure}
     \hfill
     \begin{subfigure}[b]{0.45\columnwidth}
         \centering
         \includegraphics[width=\textwidth]{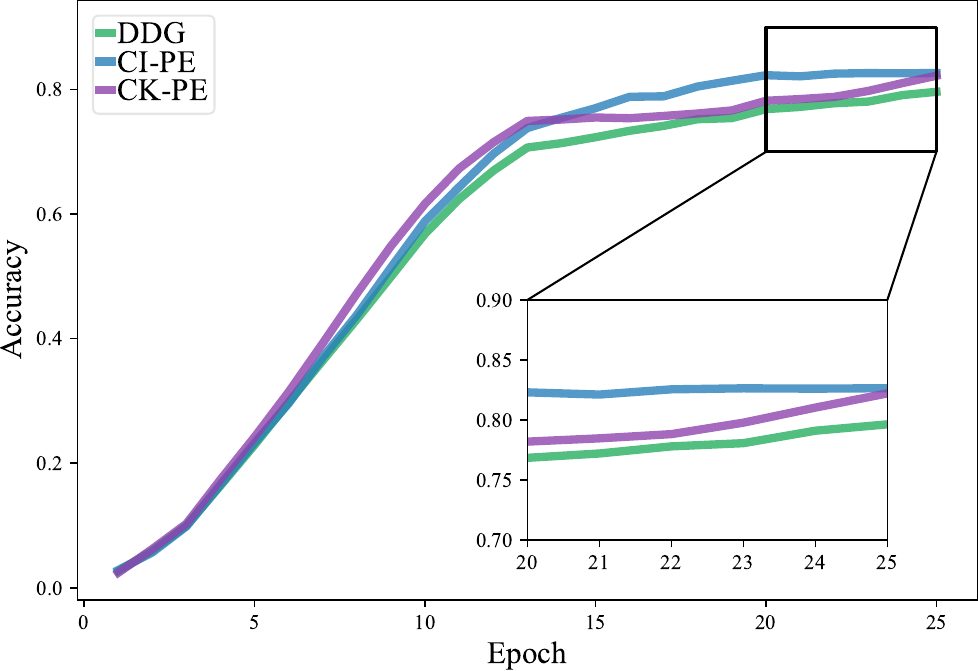}
        \vspace{-0.5cm}
         \caption{Domain classification accuracy of dynamic features}
         \label{fig:domain_accuracy_dynamic}
     \end{subfigure}\\
     \vspace{-0.3cm}
        \caption{Accuracy curves of training domain classification task on the features provided by the (a) static component or (b) dynamic component.  }
        \label{fig:domain_accuracy_curves_all}
        \vspace{-0.5cm}
\end{figure}
To further analyze our method, we utilize t-SNE~\cite{van2008visualizing} to visualize the static and dynamic components of different methods (i.e. DDG, CI-PE and CK-PE) trained with the \emph{Art}, \emph{Cartoon} and \emph{Photo} domains of PACS dataset, and evaluated on the unseen \emph{Sketch} domain. 

\noindent\paragraph{\textbf{t-SNE Visualization of the Static Component}}  
Figure \ref{fig:static_ddg}, \ref{fig:static_cipe} and \ref{fig:static_ckpe} provide visualizations of the features of the static component generated by models trained using DDG, CI-PE and CK-PE, respectively. The static features provided by CI-PE and CK-PE are observed to be more center-clustered within each category across different domains, including the unseen target domain and source domains (marked by red circles). These results suggest that our PE methods enable the static component to generate more robust domain-invariant features. 
\noindent\paragraph{\textbf{t-SNE Visualization of the Dynamic Component}}
Visualizations of the dynamic coefficients generated by different models are depicted in Figure \ref{fig:dynamic_ddg}-\ref{fig:dynamic_ckpe}, revealing that the dynamic coefficients exhibit a stronger correlation with domain labels in the CI-PE and CK-PE methods. This domain cluster-centered behavior is particularly noticeable in the unseen target domain (marked by red circles). The results suggest that our PE methods can effectively enable the dynamic component to extract more adaptive domain-specific features, especially in the case of the unseen target domain data.
\noindent\paragraph{\textbf{Domain Classification Accuracy of the Static and Dynamic Components}}
We further demonstrate the disentanglement effect of our method by training domain classification tasks on the static and dynamic features provided by the DDG, CI-PE, and CK-PE model, respectively. To achieve this, we use a shallow network that takes static or dynamic features as input, and the training curves are presented in Figure \ref{fig:domain_accuracy_curves_all}. As illustrated in Figure \ref{fig:domain_accuracy_static}, the static features of CI-PE and CK-PE consistently exhibit lower domain classification accuracy than those of DDG, indicating that the domain-invariant features learned by the static component is improved through PE. On the other hand, Figure \ref{fig:domain_accuracy_dynamic} shows that the dynamic features of CI-PE and CK-PE demonstrate higher domain classification accuracy than those of DDG, suggesting that the dynamic component captures more domain-specific information via PE. 
These evidences demonstrate that our PE method can effectively facilitate the disentanglement of domain-invariant and -specific features for DDG.

\begin{table}[t]
  \centering
  \caption{Comparisons of different methods on PACS dataset for single-source DG task. A, C, P, S denote the abbreviated names of Art, Cartoon, Painting and Sketch domains.
  The first row denotes the source domain, and the second row with abbreviated name denotes the target domains.}
   \vspace{-10pt}
  \label{tab:ablation_pacs_sdg}%
  \setlength\tabcolsep{9pt}\scalebox{0.75}{
    \begin{tabular}{c|cccc|c}
    \toprule
    \multirow{2}[2]{*}{Method} & Art   & Cartoon & Photo & Sketch & \multirow{2}[2]{*}{Avg.} \\
          & C, P, S & A, P, S & A, C, S & A, C, P &  \\
    \midrule
    DeepAll~\cite{meng2022attention} & 69.90 & 73.90 & 41.60 & \textbf{48.30} & 58.60 \\
    RSC \cite{huang2020rsc} & 70.70 & 75.10 & 41.60 & \uline{47.30} & 58.70 \\
    SelfReg \cite{kim2021selfreg} & 72.60 & 76.60 & 43.50 & 45.80 & 59.60 \\
    \midrule
    DDG \cite{sun2022ddg} & 70.74 & 77.55 & 43.58 & 36.52 & 57.10 \\
    DDG w/ CI-PE (Ours) & \uline{75.16} & \uline{77.65} & \textbf{48.50} & 44.08 & \uline{61.35} \\
    DDG w/ CK-PE (Ours) & \textbf{75.60} & \textbf{78.76} & \uline{47.64} & 46.31 & \textbf{62.08} \\
    \bottomrule
    \end{tabular}%
  }
  \vspace{-0.1cm}
\end{table}%
\subsection{Extension to Single-Source DG}\label{sec:singleDG}
We also investigate the effectiveness of PE on single-source DG task. Compared to conventional DG, single-source DG is a more challenging task because the model is trained on a single source domain and evaluated on multiple target domains. The experiments are performed on the PACS dataset, with the backbone as ResNet-50, and the training settings keep consistent with previous experiments. The comparison results with other single-source DG methods are reported in Table \ref{tab:ablation_pacs_sdg}. We can see that our method can improve the baseline (DDG) from 57.10\% to 61.35\% (CI-PE) and 62.08\% (CK- PE), which are superior to the performance of recent DG methods, e.g., RSC \cite{huang2020rsc} and SelfReg \cite{kim2021selfreg}. Our PE method has been demonstrated to be effective not just for conventional DG task but also for single-source DG task, highlighting its remarkable versatility in enhancing the generalizability of dynamic networks.

\section{Conclusion}
In this paper, we attempt to push the limits of \emph{Dynamic Domain Generalization (DDG)} from an optimization perspective that disentangles the domain-invariant and -specific features more thoroughly. To this end, based on the existing arts, we propose a novel \emph{parameter exchange} method to perturb and reassemble the combination between the static and dynamic components at the instance-level and kernel-level, respectively. The static component and the dynamic component are disentangled to obtain a more robust domain-invariant features and a more adaptive domain-specific features respectively, forming a more generalizable feature representation for each instance.  Our method has been extensively evaluated and demonstrated to be effective for DG task. Moreover, our method is plug-and-play for multiple dynamic networks, which can provide new insights to the community of dynamic networks.

\begin{acks}
This work was supported by the Fujian Provincial Natural Science Foundation (No. 2022J05135), the University-Industry Cooperation Project of Fujian Provincial Department of Science and Technology (No. 2020H6005), and the National Natural Science Foundation of China (No. U21A20471).
\end{acks}

\clearpage
\balance
\bibliographystyle{ACM-Reference-Format}
\bibliography{paper}

\end{document}